\documentclass[twoside,12pt,dina4]{article}
\catcode`\@=11
\dimen\footins=24cm

\reversemarginpar
\ifcase\@ptsize
 \font\tenmsx=msam10
 \font\sevenmsx=msam7
 \font\fivemsx=msam5
 \font\tenmsy=msbm10
 \font\sevenmsy=msbm7
 \font\fivemsy=msbm5
\or
 \font\tenmsx=msam10 scaled \magstephalf
 \font\sevenmsx=msam8
 \font\fivemsx=msam6
 \font\tenmsy=msbm10 scaled \magstephalf
 \font\sevenmsy=msbm8
 \font\fivemsy=msbm6
\or
 \font\tenmsx=msam10 scaled \magstep1
 \font\sevenmsx=msam8
 \font\fivemsx=msam6
 \font\tenmsy=msbm10 scaled \magstep1
 \font\sevenmsy=msbm8
 \font\fivemsy=msbm6
\fi
 
\newfam\msxfam
\newfam\msyfam
\textfont\msxfam=\tenmsx  \scriptfont\msxfam=\sevenmsx
  \scriptscriptfont\msxfam=\fivemsx
\textfont\msyfam=\tenmsy  \scriptfont\msyfam=\sevenmsy
  \scriptscriptfont\msyfam=\fivemsy

\def\hexnumber@#1{\ifnum#1<10 \number#1\else
 \ifnum#1=10 A\else\ifnum#1=11 B\else\ifnum#1=12 C\else
 \ifnum#1=13 D\else\ifnum#1=14 E\else\ifnum#1=15 F\fi\fi\fi\fi\fi\fi\fi}

\def\msx@{\hexnumber@\msxfam}
\def\msy@{\hexnumber@\msyfam}
\mathchardef\boxdot="2\msx@00
\mathchardef\boxplus="2\msx@01
\mathchardef\boxtimes="2\msx@02
\mathchardef\square="0\msx@03
\mathchardef\blacksquare="0\msx@04
\mathchardef\centerdot="2\msx@05
\mathchardef\lozenge="0\msx@06
\mathchardef\blacklozenge="0\msx@07
\mathchardef\circlearrowright="3\msx@08
\mathchardef\circlearrowleft="3\msx@09
\mathchardef\rightleftharpoons="3\msx@0A
\mathchardef\leftrightharpoons="3\msx@0B
\mathchardef\boxminus="2\msx@0C
\mathchardef\Vdash="3\msx@0D
\mathchardef\Vvdash="3\msx@0E
\mathchardef\vDash="3\msx@0F
\mathchardef\twoheadrightarrow="3\msx@10
\mathchardef\twoheadleftarrow="3\msx@11
\mathchardef\leftleftarrows="3\msx@12
\mathchardef\rightrightarrows="3\msx@13
\mathchardef\upuparrows="3\msx@14
\mathchardef\downdownarrows="3\msx@15
\mathchardef\upharpoonright="3\msx@16

\mathchardef\downharpoonright="3\msx@17
\mathchardef\upharpoonleft="3\msx@18
\mathchardef\downharpoonleft="3\msx@19
\mathchardef\rightarrowtail="3\msx@1A
\mathchardef\leftarrowtail="3\msx@1B
\mathchardef\leftrightarrows="3\msx@1C
\mathchardef\rightleftarrows="3\msx@1D
\mathchardef\Lsh="3\msx@1E
\mathchardef\Rsh="3\msx@1F
\mathchardef\rightsquigarrow="3\msx@20
\mathchardef\leftrightsquigarrow="3\msx@21
\mathchardef\looparrowleft="3\msx@22
\mathchardef\looparrowright="3\msx@23
\mathchardef\circeq="3\msx@24
\mathchardef\succsim="3\msx@25
\mathchardef\gtrsim="3\msx@26
\mathchardef\gtrapprox="3\msx@27
\mathchardef\multimap="3\msx@28
\mathchardef\therefore="3\msx@29
\mathchardef\because="3\msx@2A
\mathchardef\doteqdot="3\msx@2B

\mathchardef\triangleq="3\msx@2C
\mathchardef\precsim="3\msx@2D
\mathchardef\lesssim="3\msx@2E
\mathchardef\lessapprox="3\msx@2F
\mathchardef\eqslantless="3\msx@30
\mathchardef\eqslantgtr="3\msx@31
\mathchardef\curlyeqprec="3\msx@32
\mathchardef\curlyeqsucc="3\msx@33
\mathchardef\preccurlyeq="3\msx@34
\mathchardef\leqq="3\msx@35
\mathchardef\leqslant="3\msx@36
\mathchardef\lessgtr="3\msx@37
\mathchardef\backprime="0\msx@38
\mathchardef\risingdotseq="3\msx@3A
\mathchardef\fallingdotseq="3\msx@3B
\mathchardef\succcurlyeq="3\msx@3C
\mathchardef\geqq="3\msx@3D
\mathchardef\geqslant="3\msx@3E
\mathchardef\gtrless="3\msx@3F
\mathchardef\sqsubset="3\msx@40
\mathchardef\sqsupset="3\msx@41
\mathchardef\trianglerighteq="3\msx@44
\mathchardef\trianglelefteq="3\msx@45
\mathchardef\bigstar="0\msx@46
\mathchardef\between="3\msx@47
\mathchardef\blacktriangledown="0\msx@48
\mathchardef\blacktriangleright="3\msx@49
\mathchardef\blacktriangleleft="3\msx@4A
\mathchardef\blacktriangle="0\msx@4E
\mathchardef\triangledown="0\msx@4F
\mathchardef\eqcirc="3\msx@50
\mathchardef\lesseqgtr="3\msx@51
\mathchardef\gtreqless="3\msx@52
\mathchardef\lesseqqgtr="3\msx@53
\mathchardef\gtreqqless="3\msx@54
\mathchardef\Rrightarrow="3\msx@56
\mathchardef\Lleftarrow="3\msx@57
\mathchardef\veebar="2\msx@59
\mathchardef\barwedge="2\msx@5A
\mathchardef\doublebarwedge="2\msx@5B
\mathchardef\angle="0\msx@5C
\mathchardef\measuredangle="0\msx@5D
\mathchardef\sphericalangle="0\msx@5E
\mathchardef\varpropto="3\msx@5F
\mathchardef\smallsmile="3\msx@60
\mathchardef\smallfrown="3\msx@61
\mathchardef\Subset="3\msx@62
\mathchardef\Supset="3\msx@63
\mathchardef\Cup="2\msx@64

\mathchardef\Cap="2\msx@65

\mathchardef\curlywedge="2\msx@66
\mathchardef\curlyvee="2\msx@67
\mathchardef\leftthreetimes="2\msx@68
\mathchardef\rightthreetimes="2\msx@69
\mathchardef\subseteqq="3\msx@6A
\mathchardef\supseteqq="3\msx@6B
\mathchardef\bumpeq="3\msx@6C
\mathchardef\Bumpeq="3\msx@6D
\mathchardef\lll="3\msx@6E

\mathchardef\ggg="3\msx@6F

\mathchardef\circledS="0\msx@73
\mathchardef\pitchfork="3\msx@74
\mathchardef\dotplus="2\msx@75
\mathchardef\backsim="3\msx@76
\mathchardef\backsimeq="3\msx@77
\mathchardef\complement="0\msx@7B
\mathchardef\intercal="2\msx@7C
\mathchardef\circledcirc="2\msx@7D
\mathchardef\circledast="2\msx@7E
\mathchardef\circleddash="2\msx@7F
\def\ulcorner{\delimiter"4\msx@70\msx@70 }
\def\urcorner{\delimiter"5\msx@71\msx@71 }
\def\llcorner{\delimiter"4\msx@78\msx@78 }
\def\lrcorner{\delimiter"5\msx@79\msx@79 }
\def\yen{\mathhexbox\msx@55 }
\def\checkmark{\mathhexbox\msx@58 }
\def\circledR{\mathhexbox\msx@72 }
\def\maltese{\mathhexbox\msx@7A }
\mathchardef\lvertneqq="3\msy@00
\mathchardef\gvertneqq="3\msy@01
\mathchardef\nleq="3\msy@02
\mathchardef\ngeq="3\msy@03
\mathchardef\nless="3\msy@04
\mathchardef\ngtr="3\msy@05
\mathchardef\nprec="3\msy@06
\mathchardef\nsucc="3\msy@07
\mathchardef\lneqq="3\msy@08
\mathchardef\gneqq="3\msy@09
\mathchardef\nleqslant="3\msy@0A
\mathchardef\ngeqslant="3\msy@0B
\mathchardef\lneq="3\msy@0C
\mathchardef\gneq="3\msy@0D
\mathchardef\npreceq="3\msy@0E
\mathchardef\nsucceq="3\msy@0F
\mathchardef\precnsim="3\msy@10
\mathchardef\succnsim="3\msy@11
\mathchardef\lnsim="3\msy@12
\mathchardef\gnsim="3\msy@13
\mathchardef\nleqq="3\msy@14
\mathchardef\ngeqq="3\msy@15
\mathchardef\precneqq="3\msy@16
\mathchardef\succneqq="3\msy@17
\mathchardef\precnapprox="3\msy@18
\mathchardef\succnapprox="3\msy@19
\mathchardef\lnapprox="3\msy@1A
\mathchardef\gnapprox="3\msy@1B
\mathchardef\nsim="3\msy@1C
\mathchardef\napprox="3\msy@1D
\mathchardef\nsubseteqq="3\msy@22
\mathchardef\nsupseteqq="3\msy@23
\mathchardef\subsetneqq="3\msy@24
\mathchardef\supsetneqq="3\msy@25
\mathchardef\subsetneq="3\msy@28
\mathchardef\supsetneq="3\msy@29
\mathchardef\nsubseteq="3\msy@2A
\mathchardef\nsupseteq="3\msy@2B
\mathchardef\nparallel="3\msy@2C
\mathchardef\nmid="3\msy@2D
\mathchardef\nshortmid="3\msy@2E
\mathchardef\nshortparallel="3\msy@2F
\mathchardef\nvdash="3\msy@30
\mathchardef\nVdash="3\msy@31
\mathchardef\nvDash="3\msy@32
\mathchardef\nVDash="3\msy@33
\mathchardef\ntrianglerighteq="3\msy@34
\mathchardef\ntrianglelefteq="3\msy@35
\mathchardef\ntriangleleft="3\msy@36
\mathchardef\ntriangleright="3\msy@37
\mathchardef\nleftarrow="3\msy@38
\mathchardef\nrightarrow="3\msy@39
\mathchardef\nLeftarrow="3\msy@3A
\mathchardef\nRightarrow="3\msy@3B
\mathchardef\nLeftrightarrow="3\msy@3C
\mathchardef\nleftrightarrow="3\msy@3D
\mathchardef\divideontimes="2\msy@3E
\mathchardef\varnothing="0\msy@3F
\mathchardef\nexists="0\msy@40
\mathchardef\mho="0\msy@66
\mathchardef\thorn="0\msy@67
\mathchardef\beth="0\msy@69
\mathchardef\gimel="0\msy@6A
\mathchardef\daleth="0\msy@6B
\mathchardef\lessdot="3\msy@6C
\mathchardef\gtrdot="3\msy@6D
\mathchardef\ltimes="2\msy@6E
\mathchardef\rtimes="2\msy@6F
\mathchardef\shortmid="3\msy@70
\mathchardef\shortparallel="3\msy@71
\mathchardef\smallsetminus="2\msy@72
\mathchardef\thicksim="3\msy@73
\mathchardef\thickapprox="3\msy@74
\mathchardef\approxeq="3\msy@75
\mathchardef\succapprox="3\msy@76
\mathchardef\precapprox="3\msy@77
\mathchardef\curvearrowleft="3\msy@78
\mathchardef\curvearrowright="3\msy@79
\mathchardef\digamma="0\msy@7A
\mathchardef\varkappa="0\msy@7B
\mathchardef\hslash="0\msy@7D
\mathchardef\hbar="0\msy@7E
\mathchardef\backepsilon="3\msy@7F
\def\Bbb{\ifmmode\let\next\Bbb@\else
 \def\next{\errmessage{Use \string\Bbb\space only in math mode}}\fi\next}
\def\Bbb@#1{{\Bbb@@{#1}}}
\def\Bbb@@#1{\fam\msyfam#1}

\catcode`\@=12
\usepackage{times}

\catcode`\@=11
\font\sixsf  = cmss10 scaled 600
\font\fvesf  = cmss10 scaled 500
\font\egtsf  = cmss10 scaled 800

\def\xiipt{\textfont\z@\twlrm 
  \scriptfont\z@\egtrm \scriptscriptfont\z@\sixrm
\textfont\@ne\twlmi \scriptfont\@ne\egtmi \scriptscriptfont\@ne\sixmi
\textfont\tw@\twlsy \scriptfont\tw@\egtsy \scriptscriptfont\tw@\sixsy
\textfont\thr@@\tenex \scriptfont\thr@@\tenex \scriptscriptfont\thr@@\tenex
\def\unboldmath{\everymath{}\everydisplay{}\@nomath\unboldmath
          \textfont\@ne\twlmi 
          \textfont\tw@\twlsy \textfont\lyfam\twlly
          \@boldfalse}\@boldfalse
\def\boldmath{\@ifundefined{twlmib}{\global\font\twlmib\@mbi\@magscale1\global
        \font\twlsyb\@mbsy \@magscale1\global\font
         \twllyb\@lasyb\@magscale1\relax\@addfontinfo\@xiipt
              {\def\boldmath{\everymath
                {\mit}\everydisplay{\mit}\@prtct\@nomathbold
                \textfont\@ne\twlmib \textfont\tw@\twlsyb 
                \textfont\lyfam\twllyb\@prtct\@boldtrue}}}{}\@xiipt\boldmath}%
\def\prm{\fam\z@\twlrm}%
\def\pit{\fam\itfam\twlit}\textfont\itfam\twlit \scriptfont\itfam\egtit
   \scriptscriptfont\itfam\sevit
\def\psl{\fam\slfam\twlsl}\textfont\slfam\twlsl 
     \scriptfont\slfam\tensl \scriptscriptfont\slfam\tensl
\def\pbf{\fam\bffam\twlbf}\textfont\bffam\twlbf 
   \scriptfont\bffam\ninbf \scriptscriptfont\bffam\ninbf 
\def\ptt{\fam\ttfam\twltt}\textfont\ttfam\twltt
   \scriptfont\ttfam\nintt \scriptscriptfont\ttfam\nintt 
\def\psf{\fam\sffam\twlsf}\textfont\sffam\twlsf
    \scriptfont\sffam\egtsf \scriptscriptfont\sffam\sixsf
\def\psc{\@getfont\psc\scfam\@xiipt{\@mcsc\@magscale1}}%
\def\ly{\fam\lyfam\twlly}\textfont\lyfam\twlly 
   \scriptfont\lyfam\egtly \scriptscriptfont\lyfam\sixly
 \@setstrut \rm}

\def\xxvpt{\textfont\z@\twfvrm 
  \scriptfont\z@\twtyrm \scriptscriptfont\z@\svtnrm
\textfont\@ne\twtymi \scriptfont\@ne\twtymi \scriptscriptfont\@ne\svtnmi
\textfont\tw@\twtysy \scriptfont\tw@\twtysy \scriptscriptfont\tw@\svtnsy
\textfont\thr@@\tenex \scriptfont\thr@@\tenex \scriptscriptfont\thr@@\tenex
\def\unboldmath{\everymath{}\everydisplay{}\@nomath\unboldmath
        \textfont\@ne\twtymi \textfont\tw@\twtysy \textfont\lyfam\twtyly
        \@boldfalse}\@boldfalse
\def\boldmath{\@subfont\boldmath\unboldmath}%
\def\prm{\fam\z@\twfvrm}%
\def\pit{\fam\z@\twfvit}%
\def\psl{\@subfont\sl\rm}%
\def\pbf{\@getfont\pbf\bffam\@xxvpt{cmbx10\@magscale5}}%
\def\ptt{\@subfont\tt\rm}%
\def\psf{\@subfont\sf\rm}%
\def\psc{\@subfont\sc\rm}%
\def\ly{\fam\lyfam\twtyly}\textfont\lyfam\twtyly 
   \scriptfont\lyfam\twtyly \scriptscriptfont\lyfam\svtnly 
\@setstrut \rm}

\catcode`\@=12

\long\def\math#1{\relax\ifmmode{#1}\else$#1 $\fi}
\long\def\nomath#1{\relax\ifmmode\mbox{#1}\else{#1}\fi}

\catcode`\@=11

\long\def\@reargdefmath#1[#2]#3{\@reargdef#1[#2]{\math{#3}}}

\long\def\@argdefmath#1[#2]#3{\@ifdefinable #1{\@reargdefmath#1[#2]{#3}}}

\def\mathcommand#1{\@ifnextchar [{\@argdefmath#1}{\@argdefmath#1[0]}}

\def\remathcommand#1{\edef\@tempa{\expandafter\@cdr\string
  #1\@nil}\@ifundefined{\@tempa}{\@latexerr{\string#1\space undefined}\@ehc
    }{}\@ifnextchar [{\@reargdefmath#1}{\@reargdefmath#1[0]}}

\catcode`\@=12

\def\mathapplycommand      
#1#2{\mathcommand{#1}[1]{{#2}({##1})}}
\def\mathdoubleapplycommand
#1#2{\mathcommand{#1}[2]{{#2}({##1},{##2})}}

\let\mathoverline=\overline
\remathcommand\overline[1]{\mathoverline{#1}}

\newcommand\mycomment[1]{{\iffalse{#1}\fi}}

\newlength{\disp}
\newlength{\textwidthminustwomm}
\newlength{\textwidthminushalfcm}
\newlength{\textwidthminusonecm}
\newlength{\textwidthminustwocm}
\newlength{\textwidthminusfourcm}
\newlength{\textwidthminussixcm}
\newcommand\notop   {\vspace{-1.0\topsep}}
\newcommand\yestop  {\vspace{\topsep}}

\newcommand\bigmath[1]{\ \math{#1}\ }

\newcommand\linenomath[1]{\mbox{}\hfill{#1}\hfill\mbox{}\\}

\newcommand\LINEnomath[1]{\mbox{}\hfill{#1}\hfill\mbox{}}

\newbox\yinibox
\def\initial#1{\par
\noindent
\setbox\yinibox\hbox{#1}%
\setbox\yinibox\vbox to \ht\yinibox{\offinterlineskip\hbox{{#1}~~}\vss}%
\leavevmode
\hangindent=\wd\yinibox
\hangafter=-2
\hbox to0pt{\hskip-\hangindent\box\yinibox\hfill}\ignorespaces}

\mathcommand\inparentheses[1]{\left(\begin{array}[c]{l}#1\end{array}\right)}
\mathcommand\inparenthesestight
[1]{\left(\begin{array}{@{}l@{}}#1\end{array}\right)}
\mathcommand\inparenthesesinline[1]{(\ #1\ )}
\mathcommand\inpit[1]{(#1)}
\mathcommand\inparenthesesinlinetight[1]{\inpit{#1}}
\mathcommand\noparenthesesoplist[1]
{{\begin{array}{ll}\relax&#1\\\end{array}}}
\mathcommand\inparenthesesoplist[1]
{{\left(\noparenthesesoplist{#1}\right)}}

\mathcommand\displayopt
[1]{\left[\begin{array}[t]{@{\,}c@{\,}}#1\end{array}\right]}

\newcommand\stopq{\math{.}\penalty-1\relax\,\,} 
\mathcommand\hastype[2]{{#1}:{#2}}
\mathcommand\tighthastype[2]{{#1}\tight:{#2}}
\mathcommand\typed[3]{#1\hastype{#2}{#3}\stopq}
\mathcommand\comma{,\ \ }
\mathcommand\setwithstart{\{\ }
 
\mathcommand\setwithmarkq[1] {\ |_{#1}\ }
\mathcommand\setwithstop     {\ \}}

\mathcommand\displaysetwith
[2]{{\left\{\begin{array}{@{\ \ }l@{\ \ }|@{\ \ }l@{\ \ }}#1&#2\\\end
{array}\right\}}}
\mathcommand\displayset
[1]{{\left\{\begin{array}{@{\ \ }l@{\ \ }}#1\\\end{array}\right\}}}

\newcommand\tight[1]{\math{#1}}

\newcommand\tightsetminus   {\tight\setminus  }

\newcommand\tighttimes      {\tight\times     }

\newcommand\tightcup        {\tight\cup       }

\newcommand\boldequal  {\tight=}
\newcommand\boldunequal{\tight{\not=}}
\newcommand\boldless   {\tight<}

\newcommand\Defnombox{{\rm Def}}

\mathcommand\DEF{\Defnombox\:}

\mathcommand\tightimplies    {\Rightarrow}
\mathcommand\tightantiimplies{\Leftarrow}
\mathcommand\tightequivalent {\Leftrightarrow}
\mathcommand\tightund        {\wedge}
\mathcommand\tightoder       {\vee}
\mathcommand\tightshefferstroke{|}

\newcommand \id          {{\rm id}}
\mathcommand\allall      {\mathcal{V}}
\mathcommand\allsets     {\POWER\allall}
\mathcommand\allfinitesets{\FINITE\allall}
\mathcommand\urelements  {\mathcal{U}}
\mathcommand\algebra     {\mathcal{A}}
\mathcommand\balgebra    {\mathcal{B}}
\mathcommand\calgebra    {\mathcal{C}}
\mathcommand\dalgebra    {\mathcal{D}}
\mathcommand\ialgebra    {\mathcal{I}}
\mathcommand\salgebra    {\mathcal{S}}
\mathcommand\walgebra    {\mathcal{W}}
\mathcommand\ralgebra    {\mathcal{R}}
\mathcommand\algebraprime{\mathcal{A}'}
\mathapplycommand\FLD        {\rm field}
\mathapplycommand\DOM        {\rm dom}
\mathapplycommand\RAN        {\rm ran}
\mathapplycommand\kernel     {\rm ker}
\mathapplycommand\FINITE     {{\mathfrak P}_{\scriptscriptstyle\N}}
\mathcommand\POWERSYM        {{\mathfrak P}}
\mathapplycommand\POWER      {\POWERSYM}
\mathapplycommand\posfinPOWER{\POWERSYM_{\scriptscriptstyle\posN\!}}
\mathcommand\CARD[1]         {\,|{#1}|\,}
\mathcommand\FLOOR[1]        {\ \lfloor#1\rfloor\ }
\mathcommand\CEIL[1]         {\ \lceil#1\rceil\ }
\mathcommand\reverserelation[1]{{#1}^{-1}}
\mathcommand\domres[2]{{_{#2}\tight\upharpoonleft#1}}
\mathcommand\ranres[2]{{#1\tight\upharpoonright_{#2}}}
\mathcommand\funarg[1]{(#1)}
\mathcommand\app[2]{#1(#2)}
\mathcommand\displayapp[2]{#1\inparentheses{#2}}
\mathcommand\apptotuple[2]{#1#2}
\mathcommand\relapp[2]{\langle#2\rangle#1}
\mathcommand\displayrelapp
[2]{\left\langle\begin{array}{@{}l@{}}#2\end{array}\right\rangle\!#1}
\mathcommand\relappsin[2]{\relapp{#1}{\!\{#2\}\!}}
\mathcommand\revrelapp[2]{#1\langle#2\rangle}
\mathcommand\displayrevrelapp
[2]{#1\!\left\langle\begin{array}{@{}l@{}}#2\end{array}\right\rangle}
\mathcommand\revrelappsin[2]{\revrelapp{#1}{\!\{#2\}\!}}
\mathcommand\pair[2]{(#1,#2)}
\mathcommand\displaypair[2]{\inparentheses{#1,\ \ #2}}
\mathcommand\trip[3]{(#1,#2,#3)}
\mathcommand\displaytrip[3]{\inparentheses{#1,\ \ #2,\ \ #3}}
\mathcommand\superdisplaytrip[3]{\left(\begin
{array}{l}#1,\\#2,\\#3,\\\end{array}\right)}
\mathcommand\quar[4]{(#1,#2,#3,#4)}
\mathcommand\displayquar[4]{\inparentheses{#1,\ \ #2,\ \ #3,\ \ #4}}
\mathcommand\superdisplayquar[4]{\left(\begin
{array}{l}#1,\\#2,\\#3,\\#4\\\end{array}\right)}
\mathcommand\quin[5]{(#1,#2,#3,#4,#5)}
\mathcommand\displayquin[5]{\inparentheses{#1,\ \ #2,\ \ #3,\ \ #4,\ \ #5}}
\mathcommand\sext[6]{(#1,#2,#3,#4,#5,#6)}
\mathcommand\sept[7]{(#1,#2,#3,#4,#5,#6,#7)}
\mathcommand\oct[8]{(#1,#2,#3,#4,#5,#6,#7,#8)}
\mathcommand\ninetuple[9]{(#1,#2,#3,#4,#5,#6,#7,#8,#9)}

\mathcommand\setwithq[3]     {\setwithstart#2\setwithmarkq{#1}#3\setwithstop}
\mathcommand\setofwords[1]{#1^{\textstyle\ast}}
\mathcommand\setofwordsscriptstyle[1]{#1^{\ast}}
\mathcommand\setofnonemptywords[1]{#1^+}
\mathcommand\FUNSET   [2]{{#1}\rightarrow{#2}}
\mathcommand\PARFUNSET[2]{{#1}\leadsto   {#2}}
\mathcommand\FUNDEF   [3]{\hastype{#1}{\FUNSET   {#2}{#3}}}
\mathcommand\PARFUNDEF[3]{\hastype{#1}{\PARFUNSET{#2}{#3}}}
\mathcommand\morphcolon{::}
\mathcommand\morpharrow{\rightarrow}
\mathcommand\morph  [3]{{#1}\morphcolon{#2}\morpharrow{#3}}
\newcommand\Idname{{\rm Id}}
\mathcommand\Id[1]{{\Idname_{#1}}}

\mathcommand\equiclass[2]{{#1}{[\{{#2}\}]}}
\mathcommand\aritysugarsymbol{\rightarrow}
\mathcommand\aritysugar{\ \aritysugarsymbol\ }

\mathcommand\monus{\mathchoice
{\raisebox{-.4ex}{$\dot{\raisebox{.4ex}[0ex]{$-$}}$}}
{\raisebox{-.4ex}{$\dot{\raisebox{.4ex}[0ex]{$-$}}$}}
{\raisebox{-.3ex}{$\scriptstyle\dot{\raisebox{.3ex}[0ex]
{$\scriptstyle-$}}$}}                                
{\raisebox{-.2ex}{$\scriptscriptstyle\dot{\raisebox{.2ex}[0ex]
{$\scriptscriptstyle-$}}$}}                          
}

\mathcommand\yields{\;\vdash\;}
\mathcommand\antiyields{\;\dashv\;}
\mathcommand\tightyields{\vdash}
\mathcommand\tightantiyields{\dashv}
\mathcommand\notyields{\;\nvdash\;}
\mathcommand\refltransyields{\hspace{0.5em}\raisebox{1.3ex}{\math{\scriptstyle
  \ast}}\hspace{-1.0em}\yields}
\mathcommand\transyields
{\hspace{0.5em}\raisebox{1.3ex}{$\scriptscriptstyle
+$}\hspace
{-1.0em}\yields\hspace{0.3em}}
\mathcommand\notmodels{\:\not\!\models}

\newcommand\EVALSYM{{\rm eval}}
\mathapplycommand\EVAL{\EVALSYM}
\mathcommand\V{{\rm V}}
\mathcommand\X{{\rm X}}
\newcommand\nottight{\bigmath}
\mathcommand\refltransclosureinline[1]{{#1}^\ast}
\mathcommand\VARSYM {\mathcal{V}}
\mathapplycommand\VAR{\VARSYM}
\mathcommand\VARnoparentheses[1]{\VARSYM #1}
\mathcommand\VARsingleindex[1]{\VARSYM\!_{{#1}}}

\mathcommand\N{{\bf N}}
\mathcommand\posN{\N_{\:\!\!\mbox{\sixsf+}}}
\mathcommand\ZZ{{\bf Z}}
\mathcommand\Q{{\bf Q}}
\mathcommand\ordclass{\mathcal{O}}
\mathcommand\RE{{\bf R}}
\mathcommand\posRE{\RE_{\scriptscriptstyle +}}
\mathcommand\nonnegRE{\RE_{\scriptscriptstyle\succeq0}}

\mathcommand\Qzero{{\cal Q}_0}

\mathcommand\transclosure[1]{\stackrel{\scriptscriptstyle+}{#1}}
\mathcommand\transclosureinline[1]{{#1}^{\scriptscriptstyle+}}

\mathcommand\sigarity{\alpha}
\newcommand\sig{{\rm sig}}
\mathcommand\sigsorts  {{\mathbb S}}
\mathcommand\sigfunsym {{\mathbb F}}
\mathcommand\consfunsym{{\mathbb C}\,} 
\mathcommand\deffunsym {{\mathbb N}}
\mathcommand\Ax{\mathcal{AX}}

\mathcommand\unired{\Longrightarrow}
\mathcommand\redsimple{\longrightarrow}
\mathcommand\redindex[1]{\redsimple_{_{\!#1}}}
\mathcommand\notredindex[1]{\,\,\,\not\!\!\!\!\redsimple_{_{\!#1}}}

\mathcommand\redindexn[2]{\stackrel{#1}\redsimple_{_{\!#2}}}
\mathcommand\redparaindex[1]{\redpara_{\hskip-1.5pt\scriptscriptstyle#1}}
\mathcommand\revparaindex[1]{\revpara_{\hskip-1.5pt\scriptscriptstyle#1}}
\mathcommand\simparaindex[1]{\simpara_{\hskip-1.5pt\scriptscriptstyle#1}}
\mathcommand\uniredindex[1]{\unired_{\hskip-1.5pt\scriptscriptstyle#1}}

\mathcommand\antiunired{\Longleftarrow}
\mathcommand\antired{\longleftarrow}
\mathcommand\antiuniredindex[1]{\antiunired_{_{\!#1}}}
\mathcommand\antiredindex[1]{\antired_{_{\!#1}}}
\mathcommand\notantiredindex[1]{\,\,\,\not\!\!\!\!\antired_{_{\!#1}}}

\mathcommand\antiredindexn[2]{\stackrel{#1}\antired_{_{\!#2}}}
\mathcommand\antiredparaindex
[1]{\antiredpara_{\hskip-1.25pt\scriptscriptstyle#1}}
\mathcommand\antirevparaindex[1]
{\antirevpara_{\hskip-1.25pt\scriptscriptstyle#1}}

\mathcommand\sym{\longleftrightarrow}
\mathcommand\symindex[1]{\sym_{_{\!#1}}}
\mathcommand\symindexn[2]{\stackrel{#1}{\sym_{_{\!#2}}}}

\mathcommand\refltransclosure[1]{\stackrel{\scriptstyle\ast}{#1}}
\mathcommand\congru{\refltransclosure\sym}
\mathcommand\congruindex[1]{\congru_{_{\!#1}}}

\mathcommand\footnotecongru
{\congru}
\mathcommand\footnotecongruindex[1]{\footnotecongru_{_{#1}}}
\mathcommand\notcongru{\hspace{0.6em}\not\hspace{-0.85em}\refltransclosure\sym}
\mathcommand\notcongruindex[1]{\notcongru_{_{\!#1}}}

\mathcommand\trans{\transclosure\redsimple}
\mathcommand\transindex[1]{\trans_{_{\!#1}}}

\mathcommand\antitrans{\transclosure\antired}
\mathcommand\antitransindex[1]{\antitrans_{_{\!#1}}}

\mathcommand\refltrans{\refltransclosure\redsimple}
\mathcommand\refltransindex[1]{\refltrans_{_{\!#1}}}

\mathcommand\footnoterefltrans
{\refltrans}
\mathcommand\footnoterefltransindex[1]{\footnoterefltrans_{_{#1}}}

\mathcommand\notrefltrans
            {\hspace{0.6em}\not\hspace{-0.85em}\refltransclosure\redsimple}
\mathcommand\antirefltrans{\refltransclosure\antired}
\mathcommand\antirefltransindex[1]{\antirefltrans_{_{\!#1}}}

\mathcommand\footnoteantirefltrans
{\antirefltrans}
\mathcommand\footnoteantirefltransindex[1]{\footnoteantirefltrans_{_{#1}}}

\mathcommand\downarrowindex[1]{\downarrow_{_{#1}}}

\mathcommand\notconfluindex[1]{\notconflu_{_{#1}}}

\mathcommand\footnotenotconfluindex[1]{\footnotenotconflu_{_{#1}}}

\mathcommand\downdownarrow        {\downdownarrows}
\mathcommand\downdownarrowindex[1]{\downdownarrow_{_{#1}}}

\mathcommand\reflclosure[1]{\stackrel
{\scriptscriptstyle=}
{#1}}
\mathcommand\reflclosureinline[1]{{#1}^{\scriptscriptstyle=}}
\mathcommand\onlyonce
{\reflclosure\redsimple}
\mathcommand\antionlyonce{\reflclosure\antired}
\mathcommand\onlyonceindex[1]{\onlyonce_{_{\!#1}}}
\mathcommand\antionlyonceindex[1]{\antionlyonce_{_{\!#1}}}

\mathapplycommand\nf{\rm NF}

\mathcommand\toconditiontermssub     {\twoheadrightarrow_{_{\R,\X}}}
\mathcommand\toconditiontermssubY    {\twoheadrightarrow_{_{\R,\Y}}}
\mathcommand\toconditiontermsshortsub{\twoheadrightarrow_{_{\R,\V}}}

\newcommand \Y{{\rm Y}}

\mathcommand\Xprime{{\rm X}'}
\mathcommand\Yprime{{\rm Y}'}
\mathcommand\Xprimeprime{{\rm X}''}
\newcommand \SIG     {{\rm SIG}}
\mathcommand\CONS    {\mathcal{C}}
\mathcommand\SIGsort     [1]{(\SIG     ,#1)}
\mathcommand\CONSsort    [1]{(\CONS    ,#1)}
\mathcommand\varsigmasort[1]{(\varsigma,#1)}
\mathcommand\SIGsortindex     [1]{_{\SIG ,#1}}
\mathcommand\CONSsortindex    [1]{_{\CONS,#1}}
\mathcommand\DUNNOsortindex   [1]{_{\DUNNO,#1}}
\mathcommand\varsigmasortindex[1]{_{\varsigma,#1}}
\mathcommand\Vsig      {\V \!_{\SIG }}
\mathcommand\Xsig      {\X   _{\SIG }}
\mathcommand\Vcons     {\V \!_{\CONS}}
\mathcommand\Sigmacons {\Sigmaoffont_{\CONS}}
\mathcommand\Xcons     {\X   _{\CONS}}
\mathcommand\Vconsprime{\V'\!_{\CONS}}
\mathcommand\Vsigprime {\V'\!_{\SIG }}
\mathcommand\sigV      {\Vsig\!\uplus\!\Vcons}
\mathcommand\sigVprime {\Vsigprime\!\uplus\!\Vconsprime}
\mathcommand\hiddenVindi     [2]{\V   \!_{{#1},{#2}}}
\mathcommand\hiddenXindi     [2]{\X     _{{#1},{#2}}}
\mathcommand\hiddenVprimeindi[2]{\V'\!\!_{{#1},{#2}}}

\mathapplycommand\VARCONS{\VARsingleindex\CONS}
\mathapplycommand\VARSIG {\VARsingleindex\SIG}
\mathapplycommand\VARord {\VARsingleindex\boldless}

\mathcommand\vt{\mathcal{ T}}
\mathcommand\gt{\mathcal{GT}}
\mathcommand\hiddenvtindi[2]{\mathcal{ T}\!_{#1,#2}}
\mathcommand\hiddengtindi[2]{\mathcal{GT}\!_{#1,#2}}

\mathapplycommand\hiddentv     {\mathcal{T}}
\mathapplycommand\hiddentg     {\mathcal{GT}}
\mathapplycommand\hiddentvprime{\mathcal{T}'}

\mathcommand\SUBSYM{\mathcal{SU\hspace{-.08em}B}}
\mathdoubleapplycommand\SUBST{\SUBSYM}

\mathcommand\Isubst     {{\rm CGSUB}(\V ,\cons )}
\mathcommand\Isubstprime{{\rm CGSUB}(\V',\cons')}
\mathcommand\GENSYM     {\mathcal{G\hspace{-.04em}E\hspace{-.09em}N}}
\mathcommand\GSSYM      {{\GENSYM\hspace{-.14em}}\SUBSYM}
\mathdoubleapplycommand\GS{\GSSYM}

\mathcommand\POSSYMBOL{\mathcal{P\hspace{-.08em}O\hspace{-.1em}S}}
\mathcommand\VPOSSYMBOL{\V\hspace{-.04em}\POSSYMBOL}
\mathcommand\FPOSSYMBOL{\sigfunsym\hspace{-.04em}\POSSYMBOL}
\mathapplycommand\TPOS{\POSSYMBOL}
\mathapplycommand\VPOS{\VPOSSYMBOL}
\mathapplycommand\FPOS{\FPOSSYMBOL}
\mathcommand\neitherprefixsymbol{\parallel}
\mathcommand\neitherprefix[2]{{#1}\,\neitherprefixsymbol\,{#2}}
\mathcommand\replsuffix   [2]{[\,#1\leftarrow#2\,]}      
\mathcommand\replparsuffix[3]{\replsuffix{#1}{#2\ |\ #3}}
\mathcommand\repl         [3]{#1\penalty-1\replsuffix{#2}{#3}}
\mathcommand\replpar      [4]{#1\penalty-1\replparsuffix{#2}{#3}{#4}}
\mathcommand\eqvariablessig[1]{\mbox{\rm Eq}(\sig,{#1})}

\mathcommand\eqgsig{\mbox{\rm GEq}(\sig)}
\mathcommand\direqv[1]{\mbox{\rm DEq}(#1)}

\mathapplycommand\MGU{\rm Mgu}
\mathdoubleapplycommand\minmgu{\rm mgu}
\mathdoubleapplycommand\sep{\rm Sep}
\mathdoubleapplycommand\minsep{\rm sep}
\mathapplycommand\depth{\rm depth}
\mathapplycommand\size{\rm size}

\newcommand\hiddensubST{_{_{\rm ST}}}
\mathcommand\superterm   {\rhd\hiddensubST}
\mathcommand\notsuperterm{\ntriangleright\hiddensubST}
\mathcommand\subterm     {\lhd\hiddensubST}
\mathcommand\notsubterm  {\ntriangleleft\hiddensubST}
\mathcommand\supertermeq {\trianglerighteq\hiddensubST}
\mathcommand\subtermeq   {\trianglelefteq\hiddensubST}

\mathcommand\sugarregel{l\boldequal r\rulesugar C}
\mathcommand\sugarregelindex[1]{l_{#1}\boldequal r_{#1}\rulesugar C_{#1}}
\mathcommand\sugaruncondregelindex[1]{l_{#1}\boldequal r_{#1}}
\mathcommand\ndrulewithassumption[3]{{{{\scriptstyle\lbrack}#1{\scriptstyle
\rbrack}}\atop\scriptstyle #2}\over #3}

\def\R{{\rm R}}
\mathcommand\RCONS{\R_\CONS}
\mathcommand\Rprime{{\rm R}'}
\mathcommand\RX{\R,\X}
\mathcommand\RCONSX{\RCONS,\X}
\mathcommand\DEFUNSYM{\mathcal{DEFUN}}
\mathapplycommand\DEFUN{\DEFUNSYM}

\mathcommand\ATSYM  {\mathcal{A\hspace{-.02em}T}}
\mathdoubleapplycommand\atomsof{\ATSYM}

\mathcommand\GENATSYM  {\GENSYM\hspace{-.18em}\ATSYM}
\mathdoubleapplycommand\genatomsof{\GENATSYM}

\mathcommand\CONDSYM  {\mathcal{C\hspace{-.1em}O\hspace{-.12em}N\hspace{-.12em}D}}
\mathcommand\LITSYM   {\mathcal{LI\hspace{-.04em}T}}
\mathcommand\CONDLITSYM{\CONDSYM\hspace{-.04em}\LITSYM}
\mathapplycommand\condlitwitharg{\CONDLITSYM}

\mathdoubleapplycommand\literals{\LITSYM}

\mathcommand\GENLITSYM{\GENSYM\hspace{-.08em}\LITSYM}
\mathdoubleapplycommand\genliterals{\GENLITSYM}

\mathcommand\ident[1]{\mathsf{#1}}

\newcommand\set     {\ident{set}}

\newcommand\term    {\ident{term}}

\newcommand\naturalssymbol{\ident{naturals}}
\newcommand\gensymsymbol{\ident{gensym}}
\mathcommand\mbpsymbol{\ident{m\hspace{-0.055em}b\hspace{-0.045em}p}}

\newcommand\csymbol     {\ident c}
\newcommand\esymbol     {\ident e}
\newcommand\fsymbol     {\ident f}
\newcommand\gsymbol     {\ident g}
\newcommand\hsymbol     {\ident h}
\newcommand\ksymbol     {\ident k}
\newcommand\ssymbol     {\ident s}
\newcommand\Everysymbol {\ident{Every}}
\newcommand\Permsymbol {\ident{Perm}}
\newcommand\RExistssymbol{\ident{Rexists}}
\newcommand\invertsymbol{\ident{invert}}
\newcommand\invsymbol{\ident{inv}}
\newcommand\abssymbol   {\ident{abs}}
\newcommand\cnssymbol   {\ident{cons}}
\mathcommand\cnsindexsymbol[1]{\ident{cons}_{#1}}

\newcommand\lengthsymbol{\ident{length}}
\newcommand\dlsymbol    {\ident{dl}}
\newcommand\dloncesymbol{\ident{delonce}}
\newcommand\rcsymbol    {\ident{rc}}
\newcommand\brsymbol    {\ident{br}}
\newcommand\revtailsymbol{\ident{revtail}}
\newcommand\revsymbol{\ident{rev}}
\newcommand\appendsymbol {\ident{append}}
\newcommand\zeropredicatesymbol{\ident{zerop}}
\newcommand\eqsymbol        {\ident{eq}}
\newcommand\ifthensymbol    {\mbox{\ident{If{}Then}}}
\newcommand\ifthenelsesymbol{\mbox{\ident{If{}ThenElse}}}
\mathcommand\eqindexsymbol        [1]{\eqsymbol        _{#1}}
\mathcommand\ifthenindexsymbol    [1]{\ifthensymbol    _{#1}}
\mathcommand\ifthenelseindexsymbol[1]{\ifthenelsesymbol_{#1}}
\newcommand\orsymbol    {\ident{or}}
\newcommand\andsymbol   {\ident{and}}
\newcommand\leqsymbol   {\ident{leq}}
\newcommand\lessymbol   {\ident{less}}
\newcommand\lexsymbol   {\ident{lex}}
\newcommand\acksymbol   {\ident{ack}}
\newcommand\switchsymbol{\ident{switch}}
\newcommand\swatchsymbol{\ident{swatch}}
\newcommand\diveinssymbol{\ident{div1}}
\newcommand\divzweisymbol{\ident{div2}}
\newcommand\divrestsymbol{\ident{divrest}}
\newcommand\diveinstailsymbol{\ident{div1tail}}
\newcommand\divzweitailsymbol{\ident{div2tail}}

\newcommand\turingmachinesymbol{\ident T}
\newcommand\terminatespsymbol  {\ident{terminatesp}}
\newcommand\statesymbol        {\ident{state}}
\newcommand\cmdsymbol          {\ident{cmd}}
\newcommand\nthsymbol          {\ident{nth}}
\newcommand\doublesymbol       {\ident{double}}

\newcommand\ppsymbol           {\ident{p}}
\newcommand\qpsymbol           {\ident{q}}
\newcommand\Epsymbol           {\ident{E}}
\newcommand\Ppsymbol           {\ident{P}}
\newcommand\Qpsymbol           {\ident{Q}}
\newcommand\Marriessymbol      {\ident{Marries}}
\newcommand\Lovessymbol        {\ident{Loves}}
\newcommand\StolenBysymbol     {\ident{StolenBy}}
\newcommand\Humansymbol        {\ident{Human}}
\newcommand\Evensymbol         {\ident{Even}}
\newcommand\Oddsymbol          {\ident{Odd}}
\newcommand\Primesymbol        {\ident{Prime}}
\newcommand\EveryPairsymbol   {\ident{EveryPair}}
\newcommand\Givesymbol         {\ident{Give}}
\newcommand\Fathersymbol       {\ident{Father}}
\newcommand\Elephantpsymbol    {\ident{Elephant}}
\newcommand\Flowerpsymbol    {\ident{Flower}}
\newcommand\Germanpsymbol      {\ident{German}}
\newcommand\Bicyclepsymbol     {\ident{Bicycle}}
\newcommand\Hugepsymbol        {\ident{Huge}}
\newcommand\Animalpsymbol      {\ident{Animal}}
\newcommand\Malepsymbol        {\ident{Male}}
\newcommand\Boypsymbol        {\ident{Boy}}
\newcommand\Girlpsymbol        {\ident{Girl}}
\newcommand\Femalepsymbol      {\ident{Female}}
\newcommand\Roundpsymbol       {\ident{Round}}
\newcommand\Quadrangularpsymbol{\ident{Quadrangular}}
\newcommand\Metpsymbol         {\ident{Met}}
\newcommand\Bishopsymbol       {\ident{Bishop}}
\newcommand\mindexsymbol[1]{\existsvari w{#1}}

\newcommand\nonnegpsymbol      {\ident{nonnegp}}
\newcommand\wellsymbol         {\ident{well}}
\newcommand\welltailsymbol     {\ident{welltail}}
\newcommand\varsymbol          {\ident{var}}
\newcommand\aritysymbol        {\ident{arity}}

\newcommand\whilesymbol        {\ident{while}}

\newcommand\nullsymbol         {\ident{null}}
\newcommand\hdsymbol           {\ident{hd}}
\newcommand\tlsymbol           {\ident{tl}}
\newcommand\insymbol           {\ident{in}}
\newcommand\applysymbol        {\ident{app}}
\newcommand\termsymbol         {\ident{term}}
\mathcommand\tightim{\longrightarrow}
\mathcommand\im{\ \tightim\ }
\mathcommand\rs{\:\rulesugar\:\:}
\mathcommand\rulesugar{\longleftarrow}

\mathcommand\doublepp[1]      {\doublesymbol   \beginargs{#1}\allargs}
\mathcommand\aritypp[1]      {\aritysymbol   \beginargs{#1}\allargs}
\mathcommand\lengthpp[1]      {\lengthsymbol   \beginargs{#1}\allargs}
\mathcommand\wellpp[1]      {\wellsymbol   \beginargs{#1}\allargs}
\mathcommand\welltailpp[1]      {\welltailsymbol   \beginargs{#1}\allargs}
\mathcommand\varpp[1]      {\varsymbol   \beginargs{#1}\allargs}
\mathcommand\divrestpp[2]    {\divrestsymbol\beginargs{#1}\separgs{#2}\allargs}
\mathcommand\diveinspp[2]    {\diveinssymbol\beginargs{#1}\separgs{#2}\allargs}
\mathcommand\divzweipp[3]    {\divzweisymbol\beginargs{#1}\separgs{#2}
\separgs{#3}\allargs}
\mathcommand\diveinstailpp[4]    {\diveinstailsymbol\beginargs{#1}\separgs{#2}
\separgs{#3}\separgs{#4}\allargs}
\mathcommand\divzweitailpp[6]    {\divzweitailsymbol\beginargs{#1}\separgs{#2}
\separgs{#3}\separgs{#4}\separgs{#5}\separgs{#6}\allargs}
\mathcommand\mbppp[2]         {\mbpsymbol   \beginargs{#1}\separgs{#2}\allargs}
\mathcommand\revpp[1]     
{\revsymbol\beginargs{#1}\allargs}
\mathcommand\revppi[2]     
{\revsymbol^{#1}\beginargs{#2}\allargs}
\mathcommand\revtailpp[2]     
{\revtailsymbol\beginargs{#1}\separgs{#2}\allargs}
\mathcommand\revtailppi[3]
{\revtailsymbol^{#1}\beginargs{#2}\separgs{#3}\allargs}
\mathcommand\Permpp[2]     
{\Permsymbol\beginargs{#1}\separgs{#2}\allargs}
\mathcommand\Permppi[3]
{\Permsymbol^{#1}\beginargs{#2}\separgs{#3}\allargs}
\mathcommand\appendpp[2]      
{\appendsymbol \beginargs{#1}\separgs{#2}\allargs}
\mathcommand\appendppi[3]      
{\appendsymbol^{#1}\beginargs{#2}\separgs{#3}\allargs}
\mathcommand\Everypp[2]      
{\Everysymbol \beginargs{#1}\separgs{#2}\allargs}
\mathcommand\RExistspp[1]      
{\RExistssymbol \beginargs{#1}\allargs}
\mathcommand\appendlongpp[2]      
{\appendsymbol\left(\begin{array}{@{}l@{}}{#1}\separgs\\{#2}\end{array}\right)}
\mathcommand\cnspp[2]         {\cnssymbol   \beginargs{#1}\separgs{#2}\allargs}
\mathcommand\cnsppi[3]       {\cnssymbol^{#1}\beginargs{#2}\separgs{#3}\allargs}
\mathcommand\cnsindexpp[3]
{\cnsindexsymbol{#1}\beginargs{#2}\separgs{#3}\allargs}
\mathcommand\dlpp[2]          {\dlsymbol    \beginargs{#1}\separgs{#2}\allargs}
\mathcommand\dloncepp[2]      {\dloncesymbol\beginargs{#1}\separgs{#2}\allargs}
\mathcommand\dlonceppi[3]{\dloncesymbol^{#1}\beginargs{#2}\separgs{#3}\allargs}
\mathcommand\rcpp[2]          {\rcsymbol    \beginargs{#1}\separgs{#2}\allargs}
\mathcommand\brpp[2]          {\brsymbol    \beginargs{#1}\separgs{#2}\allargs}
\mathcommand\orpp[2]          {\orsymbol    \beginargs{#1}\separgs{#2}\allargs}
\mathcommand\andpp[2]         {\andsymbol   \beginargs{#1}\separgs{#2}\allargs}
\mathcommand\shortcnspp[2]    {\csymbol     \beginargs{#1}\separgs{#2}\allargs}
\mathcommand\tightshortcnspp[2]
{\csymbol\beginargs{#1}\tightsepargs{#2}\allargs}
\mathcommand\spp[1]           {\ssymbol     \beginargs{#1}\allargs}
\mathcommand\sppiterated[2]   {\ssymbol^{#1}\beginargs{#2}\allargs}
\mathcommand\ppp[1]           {\psymbol     \beginargs{#1}\allargs}
\mathcommand\pppiterated[2]   {\psymbol^{#1}\beginargs{#2}\allargs}
\mathcommand\zeropp           {\ident 0}
\mathcommand\Julietpp         {\ident{Juliet}}
\mathcommand\Romeopp          {\ident{Romeo}}
\mathcommand\Ipp              {\ident I}
\mathcommand\onepp            {\ident1}
\mathcommand\twopp            {\ident2}
\mathcommand\threepp          {\ident3}
\mathcommand\invertpp[1]      {\invertsymbol\beginargs{#1}\allargs}
\mathcommand\invpp[1]         {\invsymbol\beginargs{#1}\allargs}
\mathcommand\abspp[1]         {\abssymbol\beginargs{#1}\allargs}
\mathcommand\naturalspp[1]    {\naturalssymbol\beginargs{#1}\allargs}
\mathcommand\gensympp[1]      {\gensymsymbol\beginargs{#1}\allargs}
\mathcommand\nilpp            {\ident{nil}}
\mathcommand\falsepp          {\ident{false}}
\mathcommand\truepp           {\ident{true}}
\mathcommand\FALSEpp          {\ident{FALSE}}
\mathcommand\TRUEpp           {\ident{TRUE}}
\mathcommand\weirdppp         {\ident{weirdp}}
\mathcommand\ambigppp         {\ident{ambigp}}
\mathcommand\zeropredicatepp[1]{\zeropredicatesymbol\beginargs{#1}\allargs}
\mathcommand\cppeins       [1]{\csymbol     \beginargs{#1}\allargs}
\mathcommand\cppzwei       [2]{\csymbol\beginargs{#1}\separgs{#2}\allargs}
\mathcommand\eppeins       [1]{\esymbol     \beginargs{#1}\allargs}
\mathcommand\fppeins       [1]{\fsymbol     \beginargs{#1}\allargs}
\mathcommand\fppeinsindex  [2]{\fsymbol_{#1}\beginargs{#2}\allargs}
\mathcommand\fppeinsiterated[2]{\fsymbol^{#1}\beginargs{#2}\allargs}
\mathcommand\gppeins       [1]{\gsymbol     \beginargs{#1}\allargs}
\mathcommand\gppzwei       [2]{\gsymbol     \beginargs{#1}\separgs{#2}\allargs}
\mathcommand\hppeins       [1]{\hsymbol     \beginargs{#1}\allargs}
\mathcommand\kppeins       [1]{\ksymbol     \beginargs{#1}\allargs}
\mathcommand\appzero          {\ident a}
\mathcommand\bppzero          {\ident b}
\mathcommand\cppzero          {\ident c}
\mathcommand\dppzero          {\ident d}
\mathcommand\eppzero          {\ident e}
\mathcommand\eqindexpp[3]{\eqindexsymbol{#1}\beginargs{#2}\separgs{#3}\allargs}
\mathcommand\ifthenindexpp
[3]{\ifthenindexsymbol{#1}\beginargs{#2}\separgs{#3}\allargs}
\mathcommand\ifthenelseindexpp
[4]{\ifthenelseindexsymbol{#1}\beginargs{#2}\separgs{#3}\separgs{#4}\allargs}
\mathcommand\eqpp[2]{\eqsymbol\beginargs{#1}\separgs{#2}\allargs}
\mathcommand\leqpp[2]{\leqsymbol\beginargs{#1}\separgs{#2}\allargs}
\mathcommand\lespp[2]{\lessymbol\beginargs{#1}\separgs{#2}\allargs}
\mathcommand\lexpp[3]{\lexsymbol\beginargs{#1}\separgs{#2}\separgs{#3}\allargs}
\mathcommand\ackpp[2]{\acksymbol\beginargs{#1}\separgs{#2}\allargs}
\mathcommand\switchpp[1]{\switchsymbol\beginargs{#1}\allargs}
\mathcommand\swatchpp[1]{\swatchsymbol\beginargs{#1}\allargs}
\mathcommand\whilepp[2]{\whilesymbol\beginargs{#1}\separgs{#2}\allargs}
\mathcommand\nullpp[1]{\nullsymbol\beginargs{#1}\allargs}
\mathcommand\nullppiterated[2]{\nullsymbol^{#1}\beginargs{#2}\allargs}
\mathcommand\hdpp[1]{\hdsymbol\beginargs{#1}\allargs}
\mathcommand\hdppiterated[2]{\hdsymbol^{#1}\beginargs{#2}\allargs}
\mathcommand\tlpp[1]{\tlsymbol\beginargs{#1}\allargs}
\mathcommand\tlppiterated[2]{\tlsymbol^{#1}\beginargs{#2}\allargs}
\mathcommand\inpp[2]{\insymbol\beginargs{#1}\separgs{#2}\allargs}
\mathcommand\inppiterated[3]{\insymbol^{#1}\beginargs{#2}\separgs{#3}\allargs}
\mathcommand\applypp[2]{\applysymbol\beginargs{#1}\separgs{#2}\allargs}
\mathcommand\applyppiterated
[3]{\applysymbol^{#1}\beginargs{#2}\separgs{#3}\allargs}
\mathcommand\termpp[2]{\termsymbol\beginargs{#1}\separgs{#2}\allargs}
\mathcommand\setpp[1]{\set\beginargs{#1}\allargs}

\mathcommand\Tpp[6]{\turingmachinesymbol\beginargs{#1}\separgs{#2}\separgs
{#3}\separgs{#4}\separgs{#5}\separgs{#6}\allargs}
\mathcommand\Tppseven[7]{\turingmachinesymbol\beginargs{#1}\separgs{#2}\separgs
{#3}\separgs{#4}\separgs{#5}\separgs{#6}\separgs{#7}\allargs}
\mathcommand\foreverppp[6]{\ident{foreverp}\beginargs{#1}\separgs{#2}\separgs
{#3}\separgs{#4}\separgs{#5}\separgs{#6}\allargs}
\mathcommand\terminatesppp[6]{\terminatespsymbol\beginargs{#1}\separgs
{#2}\separgs{#3}\separgs{#4}\separgs{#5}\separgs{#6}\allargs}
\mathcommand\terminatespppone[1]{\terminatespsymbol \beginargs{#1}\allargs}
\mathcommand\statepp
[3]{\statesymbol\beginargs{#1}\separgs{#2}\separgs{#3}\allargs}
\mathcommand\tightstatepp
[3]{\statesymbol\beginargs{#1}\tightsepargs{#2}\tightsepargs{#3}\allargs}
\mathcommand\cmdpp
[3]{\cmdsymbol  \beginargs{#1}\separgs{#2}\separgs{#3}\allargs}
\mathcommand\tightcmdpp
[3]{\cmdsymbol  \beginargs{#1}\tightsepargs{#2}\tightsepargs{#3}\allargs}
\mathcommand\stoppp           {\ident{stop}}
\mathcommand\leftpp           {\ident{left}}
\mathcommand\rightpp          {\ident{right}}
\mathcommand\nthpp         [2]{\nthsymbol  \beginargs{#1}\separgs{#2}\allargs}
\mathcommand\pppp          [1]{\ppsymbol\beginargs{#1}            \allargs}
\mathcommand\qppp          [2]{\qpsymbol\beginargs{#1}\separgs{#2}\allargs}
\mathcommand\Eppp          [1]{\Epsymbol\beginargs{#1}            \allargs}
\mathcommand\Epppzwei      [2]{\Epsymbol\beginargs{#1}\separgs{#2}\allargs}
\mathcommand\Pppp          [1]{\Ppsymbol\beginargs{#1}            \allargs}
\mathcommand\Ppppdrei      
[3]{\Ppsymbol\beginargs{#1}\separgs{#2}\separgs{#3}\allargs}
\mathcommand\Ppppvier
[4]{\Ppsymbol\beginargs{#1}\separgs{#2}\separgs{#3}\separgs{#4}\allargs}
\mathcommand\Qppp          [2]{\Qpsymbol\beginargs{#1}\separgs{#2}\allargs}
\mathcommand\Qpppeins      [1]{\Qpsymbol\beginargs{#1}\allargs}
\mathcommand\Qpppdrei      
[3]{\Qpsymbol\beginargs{#1}\separgs{#2}\separgs{#3}\allargs}
\mathcommand\Fatherpp      [2]{\Fathersymbol\beginargs{#1}\separgs{#2}\allargs}
\mathcommand\Marriespp     [2]{\Marriessymbol\beginargs{#1}\separgs{#2}\allargs}
\mathcommand\Lovespp       [2]{\Lovessymbol\beginargs{#1}\separgs{#2}\allargs}
\mathcommand\StolenBypp    [2]
{\StolenBysymbol\beginargs{#1}\separgs{#2}\allargs}
\mathcommand\Humanpp       [1]{\Humansymbol\beginargs{#1}\allargs}
\mathcommand\Evenpp        [1]{\Evensymbol\beginargs{#1}\allargs}
\mathcommand\Evenppi       [2]{\Evensymbol^{#1}\beginargs{#2}\allargs}
\mathcommand\Oddpp         [1]{\Oddsymbol\beginargs{#1}\allargs}
\mathcommand\Primepp       [1]{\Primesymbol\beginargs{#1}\allargs}
\mathcommand\EveryPairpp  [2]{\EveryPairsymbol\beginargs{#1}\separgs
{#2}\allargs}
\mathcommand\mindexppeins  [2]{\mindexsymbol{#1}\beginargs{#2}\allargs}
\mathcommand\Givepp        [3]{\Givesymbol
\beginargs{#1}\separgs{#2}\separgs{#3}\allargs}
\mathcommand\mindexppzwei  [3]{\mindexsymbol
{#1}\beginargs{#2}\separgs{#3}\allargs}
\mathcommand\mindexppdrei  [4]{\mindexsymbol
{#1}\beginargs{#2}\separgs{#3}\separgs{#4}\allargs}

\mathcommand\nonnegppp     [1]{\nonnegpsymbol\beginargs{#1}\allargs}

\mathcommand\anonymouscsymbol{c}
\mathcommand\anonymouscindexsymbol[1]{\anonymouscsymbol_{#1}}
\mathcommand\anonymousfsymbol{f}
\mathcommand\anonymouscpp
[2]{\anonymouscsymbol\beginargs{#1}\separgs\ldots\separgs{#2}\allargs}
\mathcommand\anonymouscindexpp
[3]{\anonymouscindexsymbol{#1}\beginargs{#2}\separgs\ldots\separgs{#3}\allargs}
\mathcommand\anonymousfpp
[2]{\anonymousfsymbol\beginargs{#1}\separgs\ldots\separgs{#2}\allargs}
\mathcommand\coerceindexpp[3]{[#3]_{#1}^{#2}}

\mathcommand\Elephantppp    [1]{\Elephantpsymbol\beginargs{#1}\allargs}
\mathcommand\Flowerppp      [1]{\Flowerpsymbol  \beginargs{#1}\allargs}
\mathcommand\Bicycleppp     [1]{\Bicyclepsymbol \beginargs{#1}\allargs}
\mathcommand\Germanppp      [1]{\Germanpsymbol  \beginargs{#1}\allargs}
\mathcommand\Hugeppp        [1]{\Hugepsymbol    \beginargs{#1}\allargs}
\mathcommand\Animalppp      [1]{\Animalpsymbol  \beginargs{#1}\allargs}
\mathcommand\Maleppp        [1]{\Malepsymbol    \beginargs{#1}\allargs}
\mathcommand\Boyppp         [1]{\Boypsymbol     \beginargs{#1}\allargs}
\mathcommand\Girlppp        [1]{\Girlpsymbol    \beginargs{#1}\allargs}
\mathcommand\Femaleppp      [1]{\Femalepsymbol  \beginargs{#1}\allargs}
\mathcommand\Roundppp       [1]{\Roundpsymbol   \beginargs{#1}\allargs}
\mathcommand\Bishoppp       [1]{\Bishopsymbol   \beginargs{#1}\allargs}
\mathcommand\Quadrangularppp[1]{\Quadrangularpsymbol  \beginargs{#1}\allargs}
\mathcommand\Metppp[2]{\Metpsymbol     \beginargs{#1}\separgs{#2}\allargs}

\newcommand\bound     {{\rm bound}}
\newcommand\free      {{\rm free}}

\mathcommand\Vtripleindex[3]{\V\!_{{#1},\,{#2},\,{#3}}}
\mathcommand\Vdoubleindex[2]{\V\!_{{#1},\,{#2}}}
\mathcommand\Vsingleindex[1]{\V\!_{{#1}}}

\mathcommand\Erel[1]{\Gammaoffont\!_{#1}}
\mathcommand\Urel[1]{\Deltaoffont_{#1}}

\mathcommand\theRprimefromstrongtoweak{
  \inparenthesesinlinetight{
     \domres\id{\Vwall\cup\Vsome\setminus\RAN\varsigma}
     \nottight{\nottight\uplus}
     \reverserelation\varsigma
  }
  \nottight{\circ}
  \ranres
    {\transclosureinline R}
    {\Vwall\cup\Vsome\setminus\RAN\varsigma}
  \nottight{\nottight{\nottight{\uplus}}}
  \Vsome\tighttimes\Vsall
}

\mathcommand\deltaminus{\delta^-}
\mathcommand\deltaplus{\delta^+}
\mathcommand\deltaplusplus{\delta^{+^+}}
\mathcommand\deltastar{\delta^*}
\mathcommand\deltastarstar{\delta^{*^*}}

\mathcommand\Vall     {\Vsingleindex\indexdelta         }
\mathcommand\Vwall    {\Vsingleindex\indexdeltaminu     }
\mathcommand\Vsall    {\Vsingleindex\indexdeltaplus     }
\mathcommand\Vgsome   {\Vsingleindex\indexgammaplus     }
\mathcommand\Vsome    {\Vsingleindex\indexgamma         }
\mathcommand\Vfree    {\Vsingleindex\indexfree          }
\mathcommand\Vbound   {\Vsingleindex\indexbound         }
\mathcommand\Vsomesall{\Vsingleindex\indexgammadeltaplus}

\mathapplycommand\VARall      {\VARsingleindex\indexdelta         }
\mathapplycommand\VARwall     {\VARsingleindex\indexdeltaminu     }
\mathapplycommand\VARsall     {\VARsingleindex\indexdeltaplus     }
\mathapplycommand\VARgsome    {\VARsingleindex\indexgammaplus     }
\mathapplycommand\VARsome     {\VARsingleindex\indexgamma         }
\mathapplycommand\VARfree     {\VARsingleindex\indexfree          }
\mathapplycommand\VARbound    {\VARsingleindex\indexbound         }
\mathapplycommand\VARsomesall {\VARsingleindex\indexgammadeltaplus}
\mathcommand\displayVARsall[1]{\VARsingleindex\indexdeltaplus
\!\!\!\:\left(\begin{array}{@{}c@{}}#1\end{array}\right)}

\mathcommand\rigidvari     [2]{#1_{#2}^\indexgammadeltaplus}
\mathcommand\existsvari    [2]{#1_{#2}^\indexgamma    }
\mathcommand\forallvari    [2]{#1_{#2}^\indexdelta    }
\mathcommand\freevari      [2]{#1_{#2}^\indexfree     }
\mathcommand\wforallvari   [2]{#1_{#2}^\indexdeltaminu}
\mathcommand\sforallvari   [2]{#1_{#2}^\indexdeltaplus}
\mathcommand\gexistsvari   [2]{#1_{#2}^\indexgammaplus}
\mathcommand\boundvari     [2]{#1_{#2}}
\mathcommand\vari          [2]{#1_{#2}}
\mathcommand\wforallvarilow[2]{#1_{#2}^
{\raisebox{-.82ex}{\math\indexdeltaminu}}}

\newcommand\indexhelper[1]{{\scriptscriptstyle#1\:\!\!}}
\newcommand\indexdeltaplus
{\indexhelper{\delta^{\raisebox{-.17ex}{\fvesf\hskip-0.14em +}}}}
\newcommand\indexdeltaminu
{\indexhelper{\delta^{\mbox{\fvesf\hskip-0.14em\rule[.2ex]{.7em}{.15ex}}}}}
\newcommand\indexgammaplus
{\indexhelper{\gamma^{\mbox{\fvesf\hskip-0.14em +}}}}
\newcommand\indexgammadeltaplus
{\indexhelper{\gamma\delta^{\raisebox{-.17ex}{\fvesf\hskip-0.14em +}}}}

\newcommand\indexdelta{\indexhelper\delta}
\newcommand\indexgamma{\indexhelper\gamma}
\newcommand\indexfree
{{\scriptscriptstyle\free}}
\newcommand\indexbound
{{\scriptscriptstyle\bound}}

\newcommand\Wellfsymb{\ident{Wellf}}
\mathapplycommand\Wellfpp{\Wellfsymb}

\usepackage{url}

\newlength{\mybibitemsep}
\newcommand\setmybibitemsep[1]{\setlength{\mybibitemsep}{#1}}

\setmybibitemsep{0.5\topsep}

\newcommand\includenetreferences{y}

\newcommand\referencessize{\normalsize}

\newcommand\mybibbaselinestretch{0.96}

\newcommand\mybibsection[1]{\section*{#1}\if 
\addcontentslineofbibsection\addcontentsline{toc}{section}{#1}\fi}

\newcommand\addcontentslineofbibsection{y}

\newcommand\mybibtitle[1]{{\em #1\@.}}
\newcommand\mybibhardnodate
[1]{{\def\myfirstarg{#1}\ifx\empty\myfirstarg{}\else#1.\ \fi}}
\newcommand\mybibsoft
[2]{{\def\myfirstarg{#1}\def\mysecondarg{#2}\ifx\empty\myfirstarg{}\else
\if y\includenetreferences\writemynetsource#1\else\if x\includenetreferences
\ifx\empty\mysecondarg\writemynetsource#1\else{}\fi\else
\if n\includenetreferences{}\else\typeout
{WARNING includenetreferences from headerbiblio.tex has silly definition}
\fi\fi\fi\fi}}

\def\writemynetsource[#1,#2,#3,#4://#5]{{\tt\sloppy\ 
\url{#4://#5} \discretionary
{(\ignorespaces#2\,\ignorespaces#1\mbox{$\!$},\mbox{$\!$}}%
{\ignorespaces#3)\mbox{$\!$}.}%
{(\ignorespaces#2\,\ignorespaces#1\mbox{$\!$},\,\ignorespaces
#3)\mbox{$\!$}.}}}

\catcode`\@=11

\newcommand\resetlongbibstyle{%
\def\bibitem{\@ifnextchar[\@lbibitem\@bibitem}
\def\@lbibitem[##1]##2{\item[\@biblabel{##1}\hfill]\if@filesw
      {\let\protect\noexpand
       \immediate
       \write\@auxout{\string\bibcite{##2}{##1}}}\fi\ignorespaces}
\def\@bibitem##1{\item\if@filesw \immediate\write\@auxout
       {\string\bibcite{##1}{\the\value{\@listctr}}}\fi\ignorespaces}
\def\bibcite{\@newl@bel b}
\let\citation\@gobble
\let\bibdata=\@gobble
\let\bibstyle=\@gobble
\def\bibliography##1{%
  \if@filesw
    \immediate\write\@auxout{\string\bibdata{##1}}%
  \fi
  \@input@{\jobname.bbl}}
\def\bibliographystyle##1{%
  \ifx\@begindocumenthook\@undefined\else
    \expandafter\AtBeginDocument
  \fi
    {\if@filesw
       \immediate\write\@auxout{\string\bibstyle{##1}}%
     \fi}}
\def\nocite##1{\@bsphack
  \@for\@citeb:=##1\do{%
    \edef\@citeb{\expandafter\@firstofone\@citeb}%
    \if@filesw\immediate\write\@auxout{\string\citation{\@citeb}}\fi
    \@ifundefined{b@\@citeb}{\G@refundefinedtrue
        \@latex@warning{Citation `\@citeb' undefined}}{}}%
  \@esphack}
\expandafter\let\csname b@*\endcsname\@empty
\DeclareRobustCommand\cite{%
  \@ifnextchar [{\@tempswatrue\@citex}{\@tempswafalse\@citex[]}}
\DeclareRobustCommand\citet{%
  \@ifnextchar [{\@tempswatrue\@citex}{\@tempswafalse\@citex[]}}
\def\@tempswafalse{\let\if@tempswa\iffalse}
\def\@tempswatrue{\let\if@tempswa\iftrue}
\let\if@tempswa\iffalse
\def\@cite##1##2{##1\if@tempswa , ##2\fi}
\def\@citex[##1]##2{%
  \let\@citea\@empty
  \@cite{\@for\@citeb:=##2\do
    {\@citea\def\@citea{,\penalty\@m\ }%
     \edef\@citeb{\expandafter\@firstofone\@citeb}%
     \if@filesw\immediate\write\@auxout{\string\citation{\@citeb}}\fi
     \@ifundefined{b@\@citeb}{\mbox{\reset@font\bfseries ?}%
       \G@refundefinedtrue
       \@latex@warning
         {Citation `\@citeb' on page \thepage \space undefined}}%
       {\csname b@\@citeb\endcsname}}}{##1}}
\def\@biblabel##1{##1}
\def\mybibitem##1##2##3##4##5##6##7##8##9
{\item
 \if@filesw
      {\let\protect\noexpand
       \immediate
       \write\@auxout{\string\bibcite{##1}{##7\discretionary
{}{}{\,}(##3##9)}}}\fi\ignorespaces
##2 (##3##9). \mybibtitle{##4} \mybibhardnodate{##5}\mybibsoft
{##6}{##5}$\!\!$\par}
\def\thebibliography##1{\mybibsection{\refname}%
\@mkboth{\uppercase{\refname}}{\uppercase{\refname}}
\def\baselinestretch{\mybibbaselinestretch}%
\list{}{\labelwidth\z@
    \leftmargin 1.5pc
    \itemindent-\leftmargin}
    \referencessize
    \parindent\z@
    \parskip\mybibitemsep\relax
    \def\newblock{\hskip .11em plus .33em minus .07em}
    \sloppy\clubpenalty4000\widowpenalty4000
    \sfcode`\.=1000\relax}
\let\endthebibliography=\endlist
}

\newcommand\resetshortbibstyle{%
\def\bibitem{\@ifnextchar[\@lbibitem\@bibitem}
\def\@lbibitem[##1]##2{\item[\@biblabel{##1}\hfill]\if@filesw
      {\let\protect\noexpand
       \immediate
       \write\@auxout{\string\bibcite{##2}{##1}}}\fi\ignorespaces}
\def\@bibitem##1{\item\if@filesw \immediate\write\@auxout
       {\string\bibcite{##1}{\the\value{\@listctr}}}\fi\ignorespaces}
\def\bibcite{\@newl@bel b}
\let\citation\@gobble
\def\@citex[##1]##2{%
  \let\@citea\@empty
  \@cite{\@for\@citeb:=##2\do
    {\@citea\def\@citea{,\penalty\@m\ }%
     \edef\@citeb{\expandafter\@firstofone\@citeb}%
     \if@filesw\immediate\write\@auxout{\string\citation{\@citeb}}\fi
     \@ifundefined{b@\@citeb}{\mbox{\reset@font\bfseries ?}%
       \G@refundefinedtrue
       \@latex@warning
         {Citation `\@citeb' on page \thepage \space undefined}}%
       {\hbox{\csname b@\@citeb\endcsname}}}}{##1}}
\let\bibdata=\@gobble
\let\bibstyle=\@gobble
\def\bibliography##1{%
  \if@filesw
    \immediate\write\@auxout{\string\bibdata{##1}}%
  \fi
  \@input@{\jobname.bbl}}
\def\bibliographystyle##1{%
  \ifx\@begindocumenthook\@undefined\else
    \expandafter\AtBeginDocument
  \fi
    {\if@filesw
       \immediate\write\@auxout{\string\bibstyle{##1}}%
     \fi}}
\def\nocite##1{\@bsphack
  \@for\@citeb:=##1\do{%
    \edef\@citeb{\expandafter\@firstofone\@citeb}%
    \if@filesw\immediate\write\@auxout{\string\citation{\@citeb}}\fi
    \@ifundefined{b@\@citeb}{\G@refundefinedtrue
        \@latex@warning{Citation `\@citeb' undefined}}{}}%
  \@esphack}
\expandafter\let\csname b@*\endcsname\@empty
\def\@cite##1##2{[{##1\if@tempswa , ##2\fi}]}
\def\@biblabel##1{[##1]}
\DeclareRobustCommand\cite{%
  \@ifnextchar [{\@tempswatrue\@citex}{\@tempswafalse\@citex[]}}
\def\@tempswafalse{\let\if@tempswa\iffalse}
\def\@tempswatrue{\let\if@tempswa\iftrue}
\let\if@tempswa\iffalse
\def\mybibitem##1##2##3##4##5##6##7##8##9
{\bibitem[##8##9]{##1}##2 (##3). 
\mybibtitle{##4} \mybibhardnodate{##5}\mybibsoft{##6}{##5}$\!\!$\par}
\def\thebibliography##1{\mybibsection{\refname}%
\@mkboth{\uppercase{\refname}}{\uppercase{\refname}}%
\def\baselinestretch{\mybibbaselinestretch}%
\referencessize
\list
 {[\arabic{enumi}]}
 {\settowidth\labelwidth{[##1]}\leftmargin\labelwidth
 \advance\leftmargin\labelsep
 \usecounter{enumi}}
 \parskip\mybibitemsep\relax
 \def\newblock{\hskip .11em plus .33em minus .07em}
 \sloppy\clubpenalty4000\widowpenalty4000
 \sfcode`\.=1000\relax}
\let\endthebibliography=\endlist
}

\newcommand\resetnumberbibstyle{%
\def\bibitem{\@ifnextchar[\@lbibitem\@bibitem}
\def\@lbibitem[##1]##2{\item[\@biblabel{##1}\hfill]\if@filesw
      {\let\protect\noexpand
       \immediate
       \write\@auxout{\string\bibcite{##2}{##1}}}\fi\ignorespaces}
\def\@bibitem##1{\item\if@filesw \immediate\write\@auxout
       {\string\bibcite{##1}{\the\value{\@listctr}}}\fi\ignorespaces}
\def\bibcite{\@newl@bel b}
\let\citation\@gobble
\def\@citex[##1]##2{%
  \let\@citea\@empty
  \@cite{\@for\@citeb:=##2\do
    {\@citea\def\@citea{,\penalty\@m\ }%
     \edef\@citeb{\expandafter\@firstofone\@citeb}%
     \if@filesw\immediate\write\@auxout{\string\citation{\@citeb}}\fi
     \@ifundefined{b@\@citeb}{\mbox{\reset@font\bfseries ?}%
       \G@refundefinedtrue
       \@latex@warning
         {Citation `\@citeb' on page \thepage \space undefined}}%
       {\hbox{\csname b@\@citeb\endcsname}}}}{##1}}
\let\bibdata=\@gobble
\let\bibstyle=\@gobble
\def\bibliography##1{%
  \if@filesw
    \immediate\write\@auxout{\string\bibdata{##1}}%
  \fi
  \@input@{\jobname.bbl}}
\def\bibliographystyle##1{%
  \ifx\@begindocumenthook\@undefined\else
    \expandafter\AtBeginDocument
  \fi
    {\if@filesw
       \immediate\write\@auxout{\string\bibstyle{##1}}%
     \fi}}
\def\nocite##1{\@bsphack
  \@for\@citeb:=##1\do{%
    \edef\@citeb{\expandafter\@firstofone\@citeb}%
    \if@filesw\immediate\write\@auxout{\string\citation{\@citeb}}\fi
    \@ifundefined{b@\@citeb}{\G@refundefinedtrue
        \@latex@warning{Citation `\@citeb' undefined}}{}}%
  \@esphack}
\expandafter\let\csname b@*\endcsname\@empty
\def\@cite##1##2{[{##1\if@tempswa , ##2\fi}]}
\def\@biblabel##1{[##1]}
\DeclareRobustCommand\cite{%
  \@ifnextchar [{\@tempswatrue\@citex}{\@tempswafalse\@citex[]}}
\def\@tempswafalse{\let\if@tempswa\iffalse}
\def\@tempswatrue{\let\if@tempswa\iftrue}
\let\if@tempswa\iffalse
\def\mybibitem##1##2##3##4##5##6##7##8##9
{\bibitem{##1}##2 (##3). 
\mybibtitle{##4} \mybibhardnodate{##5}\mybibsoft{##6}{##5}$\!\!$\par}
\def\thebibliography##1{\mybibsection{\refname}%
\@mkboth{\uppercase{\refname}}{\uppercase{\refname}}
\def\baselinestretch{\mybibbaselinestretch}%
\referencessize
\list
 {[\arabic{enumi}]}
 {\settowidth\labelwidth{[##1]}\leftmargin\labelwidth
 \advance\leftmargin\labelsep
 \usecounter{enumi}}
 \parskip\mybibitemsep\relax
 \def\newblock{\hskip .11em plus .33em minus .07em}
 \sloppy\clubpenalty4000\widowpenalty4000
 \sfcode`\.=1000\relax}
\let\endthebibliography=\endlist
}

\newcommand\setlongbibstyle  {\AtBeginDocument{\resetlongbibstyle  }}

\catcode`\@=12
\setlongbibstyle

\usepackage{amssymb}

\mathchardef\Gammaoffont="7000
\mathchardef\Gamma="0100
\mathchardef\Deltaoffont="7001
\mathchardef\Delta="0101
\mathchardef\Thetaoffont="7002
\mathchardef\Theta="0102
\mathchardef\Lambdaoffont="7003
\mathchardef\Lambda="0103
\mathchardef\Xioffont="7004
\mathchardef\Xi="0104
\mathchardef\Pioffont="7005
\mathchardef\Pi="0105
\mathchardef\Sigmaoffont="7006
\mathchardef\Sigma="0106
\mathchardef\Upsilonoffont="7007
\mathchardef\Upsilon="0107
\mathchardef\Phioffont="7008
\mathchardef\Phi="0108
\mathchardef\Psioffont="7009
\mathchardef\Psi="0109
\mathchardef\Omegaoffont="700A
\mathchardef\Omega="010A
\mathchardef\itype="017B

\catcode`\@=11

\gdef\allowhyphens{\penalty\@M \hskip\z@skip}

\gdef\set@low@box#1{\setbox\tw@\hbox{,}\setbox\z@\hbox{#1}\dimen\z@\ht\z@
     \advance\dimen\z@ -\ht\tw@
     \setbox\z@\hbox{\lower\dimen\z@ \box\z@}\ht\z@\ht\tw@ \dp\z@\dp\tw@ }
\gdef\set@low@boxsingle#1{\setbox\tw@\hbox{\rm,}\setbox\z@\hbox{#1}\dimen\z@\ht\z@
     \advance\dimen\z@ -\ht\tw@
     \setbox\z@\hbox{\lower\dimen\z@ \box\z@}\ht\z@\ht\tw@ \dp\z@\dp\tw@ }

\gdef\@glqq{%
\ifhmode\edef\@SF{\spacefactor\the\spacefactor}%
\else\let\@SF\empty
\fi
\CheckFamily\font\fraknomath\ifSameFamily ``\relax
\else\CheckFamily\font\swab\ifSameFamily ``\relax
\else\leavevmode\set@low@box{''}\box\z@\kern-.04em\allowhyphens\@SF\relax
\fi\fi}
\gdef\glqq{\protect\@glqq\kern+.07em}
\gdef\@grqq{%
\ifhmode\edef\@SF{\spacefactor\the\spacefactor}%
\else\let\@SF\empty 
\fi 
\CheckFamily\font\fraknomath\ifSameFamily ''\relax
\else\CheckFamily\font\swab\ifSameFamily ''\relax
\else\kern+.07em``\kern.07em\@SF\relax
\fi\fi}
\gdef\grqq{\protect\@grqq}
\gdef\@glq{{\ifhmode \edef\@SF{\spacefactor\the\spacefactor}\else
     \let\@SF\empty \fi \leavevmode
     \set@low@boxsingle{'\/}\box\z@\kern-.04em\allowhyphens\@SF\relax}}
\gdef\glq{\protect\@glq\kern+.07em}
\gdef\@grq{\ifhmode \edef\@SF{\spacefactor\the\spacefactor}\else
     \let\@SF\empty \fi \kern-.0125em`\kern.07em\@SF\relax}
\gdef\grq{\protect\@grq}

\catcode`\@=12

\makeatletter
\if \@ptsize 0
   \newfont{\scriptscriptscriptgoth}{ygoth scaled 760}
   \newfont{\scriptscriptgoth}{ygoth scaled 833}
   \newfont{\scriptgoth}{ygoth scaled 912}
   \newfont{\gothnomath}{ygoth}
   \newfont{\Goth}{ygoth scaled \magstephalf}
   \newfont{\GOth}{ygoth scaled \magstep1}
   \newfont{\GOTh}{ygoth scaled \magstep2}
   \newfont{\GOTH}{ygoth scaled \magstep3}

   \newfont{\scriptscriptscriptswab}{yswab scaled 760}
   \newfont{\scriptscriptswab}{yswab scaled 833}
   \newfont{\scriptswab}{yswab scaled 912}
   \newfont{\swab}{yswab}
   \newfont{\Swab}{yswab scaled \magstephalf}
   \newfont{\SWab}{yswab scaled \magstep1}
   \newfont{\SWAb}{yswab scaled \magstep2}
   \newfont{\SWAB}{yswab scaled \magstep3}

   \newfont{\scriptscriptscriptfrak}{yfrak scaled 760}
   \newfont{\scriptscriptfrak}{yfrak scaled 833}
   \newfont{\scriptfrak}{yfrak scaled 912}
   \newfont{\fraknomath}{yfrak}
   \newfont{\Frak}{yfrak scaled \magstephalf}
   \newfont{\FRak}{yfrak scaled \magstep1}
   \newfont{\FRAk}{yfrak scaled \magstep2}
   \newfont{\FRAK}{yfrak scaled \magstep3}

   \newfont{\init}{yinit}
   \newfont{\Init}{yinit scaled \magstephalf}
   \newfont{\INit}{yinit scaled \magstep1}
   \newfont{\INIt}{yinit scaled \magstep2}
   \newfont{\INIT}{yinit scaled \magstep3}
\fi
\if \@ptsize 1
   \newfont{\scriptscriptscriptgoth}{ygoth scaled 833}
   \newfont{\scriptscriptgoth}{ygoth scaled 912}
   \newfont{\scriptgoth}{ygoth}
   \newfont{\gothnomath}{ygoth scaled \magstephalf}
   \newfont{\Goth}{ygoth scaled \magstep1}
   \newfont{\GOth}{ygoth scaled \magstep2}
   \newfont{\GOTh}{ygoth scaled \magstep3}
   \newfont{\GOTH}{ygoth scaled \magstep4}

   \newfont{\scriptscriptscriptswab}{yswab scaled 833}
   \newfont{\scriptscriptswab}{yswab scaled 912}
   \newfont{\scriptswab}{yswab}
   \newfont{\swab}{yswab scaled \magstephalf}
   \newfont{\Swab}{yswab scaled \magstep1}
   \newfont{\SWab}{yswab scaled \magstep2}
   \newfont{\SWAb}{yswab scaled \magstep3}
   \newfont{\SWAB}{yswab scaled \magstep4}

   \newfont{\scriptscriptscriptfrak}{yfrak scaled 833}
   \newfont{\scriptscriptfrak}{yfrak scaled 912}
   \newfont{\scriptfrak}{yfrak}
   \newfont{\fraknomath}{yfrak scaled \magstephalf}
   \newfont{\Frak}{yfrak scaled \magstep1}
   \newfont{\FRak}{yfrak scaled \magstep2}
   \newfont{\FRAk}{yfrak scaled \magstep3}
   \newfont{\FRAK}{yfrak scaled \magstep4}

   \newfont{\init}{yinit scaled \magstephalf}
   \newfont{\Init}{yinit scaled \magstep1}
   \newfont{\INit}{yinit scaled \magstep2}
   \newfont{\INIt}{yinit scaled \magstep3}
   \newfont{\INIT}{yinit scaled \magstep4}
\fi
\if \@ptsize 2
   \newfont{\scriptscriptscriptgoth}{ygoth scaled 912}
   \newfont{\scriptscriptgoth}{ygoth}
   \newfont{\scriptgoth}{ygoth scaled \magstephalf}
   \newfont{\gothnomath}{ygoth scaled \magstep1}
   \newfont{\Goth}{ygoth scaled \magstep2}
   \newfont{\GOth}{ygoth scaled \magstep3}
   \newfont{\GOTh}{ygoth scaled \magstep4}
   \newfont{\GOTH}{ygoth scaled \magstep5}

   \newfont{\scriptscriptscriptswab}{yswab scaled 912}
   \newfont{\scriptscriptswab}{yswab}
   \newfont{\scriptswab}{yswab scaled \magstephalf}
   \newfont{\swab}{yswab scaled \magstep1}
   \newfont{\Swab}{yswab scaled \magstep2}
   \newfont{\SWab}{yswab scaled \magstep3}
   \newfont{\SWAb}{yswab scaled \magstep4}
   \newfont{\SWAB}{yswab scaled \magstep5}

   \newfont{\scriptscriptscriptfrak}{yfrak scaled 912}
   \newfont{\scriptscriptfrak}{yfrak}
   \newfont{\scriptfrak}{yfrak scaled \magstephalf}
   \newfont{\fraknomath}{yfrak scaled \magstep1}
   \newfont{\Frak}{yfrak scaled \magstep2}
   \newfont{\FRak}{yfrak scaled \magstep3}
   \newfont{\FRAk}{yfrak scaled \magstep4}
   \newfont{\FRAK}{yfrak scaled \magstep5}

   \newfont{\init}{yinit scaled \magstep1}
   \newfont{\Init}{yinit scaled \magstep2}
   \newfont{\INit}{yinit scaled \magstep3}
   \newfont{\INIt}{yinit scaled \magstep4}
   \newfont{\INIT}{yinit scaled \magstep5}
\fi

\makeatother

\newif\ifSameFamily
\def\CheckFamily#1#2{\GetFamilyName{#1}\ArgOne
        \GetFamilyName{#2}\ArgTwo
        \ifx\ArgOne\ArgTwo\SameFamilytrue\else\SameFamilyfalse\fi}
\def\GetFamilyName#1{\edef\Tempa{#1}\def\Tempb{#1}\ifx\Tempa\Tempb
        \edef\Tempa{\fontname#1}\fi
        \edef\Tempa{\Tempa\space}%
        \expandafter\iGetFamilyName\Tempa\\}
\def\iGetFamilyName#1 #2\\#3{\def#3{#1}}
\def\DefFontName#1#2{{\escapechar-1\expandafter\expandafter\expandafter
        \iDefFontName\expandafter{\csname#2\endcsname}%
        \xdef#1{\expandafter\string\Tempa}}}
\def\iDefFontName{\def\Tempa}

\DefFontName\eccclarge{eccc1200}
\DefFontName\eccc{eccc1000}
\DefFontName\ecccsmall{eccc0900}
\DefFontName\ecccfootnotesize{eccc0800}

\newcommand\unprotectedes
{\CheckFamily\font\fraknomath\ifSameFamily\char'215\else
\CheckFamily\font\swab\ifSameFamily\char'215\else  
s\fi\fi}

\newcommand\unprotectedesi
{\CheckFamily\font\fraknomath\ifSameFamily\char'215\else
\CheckFamily\font\swab\ifSameFamily\char'215\else  
\mbox{s\hskip.05em}\fi\fi
\-\ignorespaces}

\newcommand\es  {\protect\unprotectedes}

\newcommand\esi {\protect\unprotectedesi}  

\newcommand\namefont{}

\newcommand\bernhard{Bern\-hard}

\newcommand\claus   {Clau\es}

\newcommand\peter   {Peter}

\newcommand\gramlich        {Gram\-lich}
\newcommand\gramlichname    {\bernhard\ \gramlich}

\newcommand\noether         {Noether}

\newcommand\noetherian      {\noether ian}

\newcommand\wirth           {{\namefont Wirth}}
\newcommand\wirthname       {{\namefont\claus-\peter\ \wirth}}

\newcommand\FBinf{Fachbereich Informatik}

\newcommand\cf   {cf.}

\newcommand\Eg   {E.g.}

\newcommand\uiff {\ iff\ }

\newcommand\pp   {pp.}
\newcommand\PP[2]{\pp\,\ignorespaces#1--\ignorespaces#2}

\newcommand\pnc  {posi\-tive/ne\-ga\-tive-condi\-tional}

\newcommand\resp {resp.}

\newcommand\nth[1]{\nthtinypositioner{#1}{\nthstring{#1}}}
\newcommand\nthtinypositioner[2]{#1\raisebox{0.52ex}{\tiny\hspace{0.07em}#2}}

\newcommand\modulointocountzero[2]
{\count1=#1
 \count2=#2
 \count0=\count1
 \divide  \count0 by \count2
 \multiply\count0 by-\count2
 \advance \count0 by \count1}
\newcommand\absolutevalueintocountzero[1]
{\count0=#1
 \ifnum\count0<0\multiply\count0 by -1\fi}
\newcommand\nthstring[1]
{\def\myargone{#1}\ifcat a\myargone th\else\nthstringnochar{#1}\fi}
\newcommand\nthstringnochar[1]
{\absolutevalueintocountzero{#1}%
 \modulointocountzero{\count0}{100} 
 \ifnum\count0>9\ifnum\count0<20 th\else\stupidnthstring\fi
                                   \else\stupidnthstring\fi}
\newcommand\stupidnthstring
{\modulointocountzero{\count0}{10}
 \ifnum\count0=1 \hskip-0.2em st\else
 \ifnum\count0=2 nd\else
 \ifnum\count0=3 rd\else 
                 th\fi\fi\fi}

\newcommand\writeascents
[1]{\count4=#1
\ifnum\count4<0 
-\multiply\count4 by -1\fi
\modulointocountzero{\count4}{10}
\divide\count4 by 10
\count3=\the\count0
\modulointocountzero{\count4}{10}
\divide\count4 by 10
\the\count4
.\the\count0
\the\count3
}

\newcommand\frenchnthstring[1]
{\def\myargone{#1}\ifcat a\myargone th\else\frenchnthstringnochar{#1}\fi}
\newcommand\frenchnthstringnochar[1]
{\absolutevalueintocountzero{#1}%
 \modulointocountzero{\count0}{100} 
 \ifnum\count0>9\ifnum\count0<20 th\else\frenchstupidnthstring\fi
                                   \else\frenchstupidnthstring\fi}
\newcommand\frenchstupidnthstring
{\modulointocountzero{\count0}{10}
 \ifnum\count0=1 \hskip-0.2em re\else
 \ifnum\count0=2 me\else
 \ifnum\count0=3 rd\else 
                 th\fi\fi\fi}

\newcommand\CLAM      {{\rm CL\kern-.36em\raise.39ex\hbox{\sc a}\kern-.15emM}}

\newcommand\TEXMACS   {{\sc T\kern-.1667em\lower.5ex\hbox{E}\kern-.125emX\kern-.1em\lower.5ex\hbox{\textsc{m\kern-.05ema\kern-.125emc\kern-.05ems}}}}

\newcommand\KL             {Kai\-ser\esi lau\-tern}

\newcommand\addressuniKLlongwithnl
{\FBinf,\\\uniKL,\\\plzuniKL\,\KL}

\newcommand\LNCSvol[1]
{\mbox{LNCS}\,\ignorespaces#1}

\newcommand\academicpress{Academic Press (\elsevier)}

\newcommand\elsevier{Elsevier}

\newcommand\springerverlag{Sprin\-ger}

\newcommand\lncsconf[6]{\nth{#2}\,#1\,#3, #4, \PP{#5}{#6}, \springerverlag}

\newcommand\CTRSshort{CTRS}

\newcommand\thirdCTRSninetytwo     {\lncsconf\CTRSshort{3}{1992}{\LNCSvol{656}}}

\newcommand\newspaperreference[5]
{\def\nameofjournalpress{#2}#1, #4 #5, #3\if?\nameofjournalpress
 \else, #2\fi}

\newcommand\dateinjournal[1]{}

\newcommand\journalreference[6]
{\def\nameofjournalpress{#2}#1\nolinebreak\hskip.2em%
 \dateinjournal{(#3) }{\mbox{\bf #4}}, \PP{#5}{#6}\if?\nameofjournalpress
 \else, #2\fi}

\newcommand\journalreferenceprintyear[6]
{\def\nameofjournalpress{#2}#1 
 {(#3) }{\bf #4}, \PP{#5}{#6}\if?\nameofjournalpress
 \else, #2\fi}

\newcommand\journalreferenceprintyearaspartofnumber[6]
{\def\nameofjournalpress{#2}#1 
 {#4/#3}, \PP{#5}{#6}\if?\nameofjournalpress
 \else, #2\fi}

\newcommand\jscname
{J. Symbolic Computation}

\newcommand\jscprintyear
{\journalreferenceprintyear{\jscname}\academicpress}

\newcommand\citewgctrslncs        {Wirth \& Gramlich (1993)}

\newcommand\idx[1]{\math{_{#1}}}
\newcommand\idxbox[1]{\makebox[0.4em][l]{\idx{#1}}}
\newcommand\idxbbox[1]{\makebox[0.85em][l]{\idx{#1}}}
\newcommand\idxBox[1]{\makebox[1.8em][l]{\idx{#1}}}

\newcommand{\nts}[1]{\mbox{$<$#1$>$}} 
\newcommand{\ts                }[1]{{\mbox{\tt#1}}}

\newcommand \variablename           {\nts{variable-name}}
\newcommand \constantname           {\nts{constant-name}}
\newcommand \functionname           {\nts{function-name}}

\renewcommand{\term                  }{\nts{term}}
\newcommand{\inequalityatom        }{\nts{(in-)equality-atom}}
\newcommand\negatablename{negatible}
\newcommand\Negatablename{Negatible}
\newcommand{\negatablecondition    }{\nts{\negatablename-condition}}
\newcommand{\negatableconditionlist}{\nts{\negatablename-condition-list}}
\newcommand{\defatom               }{\nts{def-atom}}
\newcommand \predatom               {\nts{predicate-atom}}
\newcommand \negatableatom          {\nts{\negatablename-atom}}
\newcommand{\basicatom             }{\nts{basic-atom}}
\newcommand{\matchatom             }{\nts{match-atom}}
\newcommand{\generalcondition      }{\nts{general-condition}}
\newcommand{\letatom               }{\nts{let-atom}}
\newcommand \generalconditionlist    {\nts{general-condition-list}}
\newcommand{\metaterm              }{\nts{meta-term}}
\newcommand \case                   {\nts{case}}
\newcommand \elsecase               {\nts{else}}
\newcommand{\caseterm              }{\nts{case-term}}
\newcommand{\negatablecase         }{\nts{\negatablename-case}}
\newcommand{\casetermwithelse      }{\nts{case-term-with-else}}
\newcommand{\ifterm                }{\nts{if-term}}
\newcommand\macrorulename  {macro-rule}
\newcommand\macrorule      {\nts{\macrorulename}}
\newcommand\macroruleintext{{\tt\macrorulename}}

\newcommand{\repsoom           }[1]{\mbox{#1$^{+}$}}
\newcommand{\reps              }[1]{\mbox{#1*}}

\newcommand{\tpar              }[1]{\ts{(}#1\ts{)}}

\newcommand{\eq                }[1]{\mbox{(=~#1)}}
\newcommand{\noteq             }[1]{\mbox{(\#~#1)}}
\newcommand\matchname      {\ts{@}}
\newcommand\doublematchname{\ts{@@}}
\newcommand\letname        {\ts{let}}
\newcommand\casename       {\ts{case}}
\newcommand\ifname         {\ts{if}}

\newcommand\elsename       {\ts{else}}

\newcommand\notname        {\ts{not}}
\newcommand\orname         {\ts{or}}
\newcommand\seqorname      {\ts{or*}}
\newcommand\andname        {\ts{and}}
\newcommand\seqandname     {\ts{and*}}
\newcommand\match      [2]{\ts{(\matchname      ~{#1}~{#2})}}
\newcommand\doublematch[2]{\ts{(\doublematchname~{#1}~{#2})}}
\newcommand\letc       [2]{\ts{(\letname        ~{#1}~{#2})}}

\newcommand{\LG                }[1]{L$_{G}$(#1)}

\newenvironment{program}{\begin{tt}\begin{tabbing} ---\=---\=---\=---\=---\=---\=---\=---\=---\=---\=---\=---\=---\= \kill 
\samepage}{\end{tabbing}\end{tt}}

\newcommand\deletename      {delete}
\newcommand\deletenameintext{{\tt\deletename}}

\newcommand\SEKIedition{%
Searchable Online Edition\\}
\begin{document}
\pagestyle{empty}
\setcounter{page}{1}
\flushbottom
\noindent
\begin{minipage}{\textwidth}
\mbox{}

\vspace{3.5em}
\begin{center}
{\LARGE\bf
   Writing \\
   Positive/Negative-Conditional Equations\\
   Conveniently
\\\mbox{}
}
{
\\\mbox{}
\\\mbox{}
\\\mbox{}
\large
Claus-Peter Wirth,
R\"{u}diger Lunde
\\\mbox{}
\\
}
{\SEKIedition
December 22, 1994
\\\mbox{}
\\\mbox{}
\\\mbox{}
SEKI-WORKING-PAPER SWP--94--04 (SFB)\\\mbox{}\\
\addressuniKLlongwithnl
\\
\mbox{}\\
\mbox{}\\

}
\end{center}
\end{minipage}

\noindent
\LINEnomath{
\begin{minipage}{\disp}
\footnotesize\renewcommand{\baselinestretch}{0.8}{\bf Abstract:}
We present a convenient notation for positive/negative-conditional
equations. The idea is to merge rules specifying the 
same function by using case-, if-, match-, and let-expressions. 
Based on the presented \macroruleintext-construct,
positive/negative-conditional 
equational specifications can be written on a higher level.
A rewrite system translates the  \macroruleintext-constructs into 
positive/negative-conditional
equations.

\end{minipage}
}

\vspace*{2.0\topsep}
\tableofcontents\nopagebreak
\vspace{\fill}

\noindent
\footnoterule
\noindent
{\footnotesize 
This research was supported by the Deutsche
Forschungsgemeinschaft, SFB 314 (D4-Projekt)

}

\cleardoublepage

\pagestyle{myheadings}%
\pagenumbering{arabic}
\setcounter{page}{1}

\vfill

\section{Introduction}
We present a \macroruleintext-construct 
for convenient specification with \pnc\ equations
as presented in \citewgctrslncs. 
Though separate equations building up the definition of one single function
are advantageous under several theoretical and practical aspects,
this separation does not correspond to the ``natural'' way of 
defining functions. 
As equational specification requires
every reduction rule to be defined explicitly, 
various repetitions of common sub-expressions occur.
In specifications with positive/negative-conditional equations, moreover,
case distinctions lead to frequent numerous
repetitions of only slightly changed left-hand sides and condition lists. 
This is rather tedious for the specifier and
a source of errors.
It also hides the actual structure of the specification. 

\yestop
\noindent 
To overcome these problems we introduce a \macroruleintext-construct
for achieving the following aims:
\begin{itemize}

\item
Concise notation: 
The specifier should be able to express the sharing of
expressions in the specification language instead of having 
to spread copies of a common sub-expression all over a function's definition.

\item 
Logical modularization: 
Reduction rules for the same function should be combined 
and structured hierarchically.

\item 
Explicit representation of case distinctive structures: 
The knowledge the specifier has in
mind should be made explicit.

\item 
Free choice of specification level: The language should also allow equational
specification without using the structural features.

\end{itemize}

\vfill
\yestop
\noindent 
To explain some ideas of our approach we will use the following rules: 

\noindent {\tt \begin{tabular}{l@{~\boldequal~}l c l}
delete x nil       & nil               &         & \\
delete x cons y k  & delete x k        & $\longleftarrow$ & x       $=$    y \\
delete x cons y k  & cons y delete x k & $\longleftarrow$ & x       $\neq$ y \\
delete x l         & l                 & $\longleftarrow$ & 
memberp x l $\not=$ true\\
\end{tabular}}
\vfill
\vfill

\pagebreak
\yestop
\yestop
\yestop
\noindent 
The main features of our \macroruleintext-construct are:
\begin{itemize}

\vfill

\item 
Conditions of equations are written as lists and 
characteristic functions as predicates.

For example the last \deletenameintext-rule above may be written

{\tt \begin{tabular}{@{}l@{~\boldequal~}l c l l}
   (delete x l) & l &$\longleftarrow$&((not (memberp x l))) & \\
   \end{tabular}}

\vfill

\item 
Contraction of right-hand sides and conditions into a new ``meta-term'',
changing the order of appearance:

Instead of 

{\tt \begin{tabular}{l@{~\boldequal~}l c l}
delete x cons y k  & delete x k        & $\longleftarrow$ & x $=$ y \\
\end{tabular}}

we write

{\tt \begin{tabular}{l@{~\boldequal~}l c l}
(delete x (cons y k))&(case &(\eq{x y}) &(delete x k))\\
\end{tabular}}

\vfill

\item 
Introduction of match-conditions \match{\tt VAR}{\tt TERM},
which bind the variables in the term {\tt TERM}
by a required match from {\tt TERM} to the value of the variable {\tt VAR}.
This has the advantage that all left-hand sides of equations
specifying the same function can be written in the same way.

The rules of our \deletenameintext-specification can now be written like this:

\noindent 
{\tt 
\begin{tabular}{@{}l l l@{}l l@{)}}
(\deletename\ x l)
&\boldequal
&(case~(
&\match{l}{nil})
&nil
\\\\(\deletename\ x l)
&\boldequal
&(case~(
&\match{l}{(cons y k)}
\\&&&\eq{x y}) 
&(\deletename\ x k)   
\\\\(\deletename\ x l)
&\boldequal
&(case~(
&\match{l}{(cons y k)} 
\\&&&\noteq{x y}) 
&(cons y (\deletename\ x k))
\\\\(\deletename\ x l)
&\boldequal
&(case~(
&(not (memberp x l))) 
&l
\\\\
\end{tabular}
}

The match-atom \match{VAR}{TERM} connects the rule's variable {\tt VAR}
with those variables that are introduced by {\tt TERM} 
and may occur to the right of the match-atom.
For avoiding reference problems,
the variables in {\tt TERM} must not occur to the left of \match{VAR}{TERM}
in the rule.

This restriction can be weakened to apply only to those
variables that are not properly 
influenced
by some let- or match-atom.
Especially {\tt VAR} may occur in {\tt TERM} and to the left of
\match{\tt VAR}{\tt TERM}. 
\Eg\ in the above rules, 
we could replace the {\tt k} with {\tt l}.
In this case, the occurrences of {\tt VAR} in the
second argument of 
\match{\tt VAR}{\tt TERM} have the same meaning as the 
occurrences of {\tt VAR} to the right of \match{\tt VAR}{\tt TERM}, 
which is different
from the meaning of {\tt VAR} in the
first argument of 
\match{\tt VAR}{\tt TERM} 
having the same meaning as the 
occurrences of {\tt VAR} to the left of \match{\tt VAR}{\tt TERM}.
Thus
(in case of {\tt VAR} occurring in {\tt TERM})
the borderline of the meaning of {\tt VAR} in the rule goes right
through the match-atom.

\vfill\pagebreak

If, however, {\tt VAR} 
does not occur in {\tt TERM}, then the meaning of \ts{VAR} to the left and to
the right of the match-atom is the same.
This persistence of the meaning of {\tt VAR} can be useful.
As an (not really convincing) example 
the first \deletenameintext-rule could be written:

\noindent 
{\tt\begin{tabular}{@{}l l l@{}l l@{)}}
(\deletename\ x l)&\boldequal&(case~(&\match{l}{nil})&l\\
\end{tabular}}

\item 
A \letname-expression \letc{TERM}{VAR}
may occur in condition lists 
and introduces {\tt VAR} as a macro for {\tt TERM}.

Each of the expressions
\match{\tt VAR\idx{2}}{\tt TERM\idx{1}}
and
\letc{\tt TERM\idx{2}}{\tt VAR\idx{1}}
binds the variables occurring in its second argument 
(\ts{TERM\idx1}, \ts{VAR\idx1}, \resp)
with the scope being the rest of the rule.
If one of these variables is already bound in the context of the expression,
then its old binding is lost in the scope of the expression.
Since this is a common source for bugs in specifications, 
the specifier should%
\footnote{For the specifier who really wants to write 
  \match{l}{(cons x l)}
  and does not want to be warned all the time, there is another
  match-atom having the form 
  \doublematch{VAR\idx{2}}{TERM\idx{1}}.
  It behaves similar to 
  \match{VAR\idx{2}}{TERM\idx{1}}
  but
  does not warn 
  if {\tt VAR\idx{2}}
  occurs in {\tt TERM\idx{1}},
  since it un-binds {\tt VAR\idx{2}} 
  before it binds the variables in {\tt TERM\idx{1}} via matching 
  {\tt TERM\idx{1}} to the old binding of {\tt VAR\idx{2}}.
}
be warned if such a re-binding occurs.
\Eg\ 
\letc{(cons x l)}{l}
re-binds {\tt l} to the term {\tt (cons x l)}
where {\tt l} refers to the old binding of {\tt l}, which is lost
for the rest of the rule.
Similarly,
if {\tt cons} is  the top symbol of {\tt l},
then \match{l}{(cons x l)}
binds {\tt x} to the first argument of {\tt l}
and re-binds {\tt l} to the second argument of the old binding of {\tt l}, 
which again is lost for the rest of the rule.

The translation into rules removes an atom 
\letc{\tt TERM\idx{2}}{\tt VAR\idx{1}}
by substituting {\tt TERM\idx{2}} for all 
occurrences of 
{\tt VAR\idx{1}} to the right of 
the atom.
Similarly, an atom
\match{\tt VAR\idx{2}}{\tt TERM\idx{1}}
is removed by 
substituting {\tt TERM\idx{1}} for all 
occurrences of 
{\tt VAR\idx{2}} to the left
and (unless {\tt VAR\idx{2}} occurs in {\tt TERM\idx{1}}) to the right of
the atom.%
\footnote{Similarly, an atom
  \doublematch{\tt VAR\idx{2}}{\tt TERM\idx{1}}
  is removed by 
  substituting {\tt TERM\idx{1}} for all {\tt VAR\idx{2}} to the left
  the atom.
}

\vfill

\item 
Equations with the same left-hand side are merged:
   \begin{program}
   ---\=---\=---\=  \kill
   (\macrorulename\ (\deletename\ x l) \\
   \> (case \\\\
   \> \> (\match{l}{nil}) \\
   \> \> nil \\\\
   \> \> (\match{l}{(cons y k)}\\
   \> \>\ (= x y)) \\
   \> \> (delete x k) \\\\
   \> \> (\match{l}{(cons y k)}\\
   \> \>\ (\# x y)) \\
   \> \> (cons y (delete x k)))) \\
   \end{program}

\vfill\pagebreak

\item 
\Negatablename\ conditions may be used in the (conjunctive) 
condition lists of \casename-with-\elsename- and \ifname-expressions.
The two latter cases of the above \macrorulename-expression can be combined
into:\notop
   {\tt \begin{tabbing}
   ------\=---\=  \kill
   \> $\vdots$ \\
   \> (\match{l}{(cons y k)})\\
   \> (if ((= x y)) \\
   \> \> (delete x k)  \\
   \> \> (cons y (delete x k))) \\
   \> $\vdots$ 
   \end{tabbing}}
\notop

\yestop
\yestop
For a condition list of length \bigmath{n+1} an ``\ifname''-expression saves 
\bigmath{n+1} condition literals
and \math{n} repetitions of  the meta-term of the else-case in the 
specification:
   {\tt \begin{tabbing}
   ---\=---\=---\=  \kill
   (\macrorulename\ l\\
   \> (\ifname\ (L\math{_0} \ldots\ L\math{_n}) \\
   \> \> r\math{_0}\\
   \> \> r\math{_1}))
   \end{tabbing}}
written in form of unstructured conditional equations is much longer:

{\tt 
\begin{tabular}{lclcl}
l
&\boldequal
&r\math{_0}
&\rulesugar 
&L\math{_0} \ldots\ L\math{_n}\\
l
&\boldequal
&r\math{_1}
&\rulesugar 
&(not L\math{_0})\\
l
&\boldequal
&r\math{_1}
&\rulesugar 
&(not L\math{_1})\\
\math\vdots
&&\math\vdots
&
&\math\vdots\\
l
&\boldequal
&r\math{_1}
&\rulesugar 
&(not L\math{_n})\\
\end{tabular}\\
}

For a \casename-with-\elsename-expression 
the saving has the complexity of the 
product of the lengths of the condition lists.
\yestop

\item
The possibility of nestling \casename- and \ifname-expressions allows 
a quadratic saving in the number of condition literals:
   {\tt \begin{tabbing}
   ---\=---\=---\=---\=  \kill
   (\macrorulename\ l\\
   \>(if (L\math{_0})\>\>\>r\math{_0} \\
   \>(if (L\math{_1})\>\>\>r\math{_1} \\
   \>\math\vdots     \>\>\>\math\vdots\\
   \>(if (L\math{_n})\>\>\>r\math{_n} \\
   \>                \>\>\>r\math{_{n+1}})\ldots)))
   \end{tabbing}}
written in form of unstructured conditional equations is much longer:

{\tt 
\begin{tabular}{lclcl}
l
&\boldequal
&r\math{_0}
&\rulesugar 
&L\math{_0}\\
l
&\boldequal
&r\math{_1}
&\rulesugar 
&(not L\math{_0}) 
L\math{_1}
\\
\math\vdots
&
&\math\vdots
&
&\math\vdots
\\
l
&\boldequal
&r\math{_n}
&\rulesugar 
&(not L\math{_0}) \ldots\ (not L\math{_{n-1}}) 
L\math{_n}
\\
l
&\boldequal
&r\math{_{n+1}}
&\rulesugar 
&(not L\math{_0}) \ldots\ (not L\math{_{n-1}}) 
(not L\math{_n})
\\
\end{tabular}
}

\vfill\pagebreak

\item
Propositional logic expressions using  ``{\tt not}'', ``\andname'', and ``\orname'' may occur
in condition lists. For example

\begin{minipage}{3cm}
{\tt \begin{tabbing}
   ---\=---\=---\=  \kill
   (\macrorulename\ l\\
   \> (\ifname\ (L\math{_0} \ldots\ L\math{_n}) \\
   \> \> r\math{_0}\\
   \> \> r\math{_1}))
   \end{tabbing}}
\end{minipage}
~~~~is equivalent to:~~~~
\begin{minipage}{7cm}
{\begin{program}
   (\macrorulename\ l\\
   \> (\casename \\
   \> \> (L\math{_0} \ldots\ L\math{_n}) \\
   \> \> r\math{_0}\\
   \> \> ((\orname\ (\notname\ L\math{_0}) \ldots\ (\notname\ L\math{_n}))) \\
   \> \> r\math{_1})) \\
   \end{program}}
\end{minipage}

Note that the positive/negative-conditional rule system,
denoted by an \orname-conditioned case contains in general more than one conditional equation
differing only in the condition part.
As we do not provide a certain order between positive/negative-conditional equations
it is of no importance in which order the arguments are supplied in the \orname-expression unless its
negation becomes relevant due to an outer ``\notname'' or a following \elsename-case.
In the denoted rule system the \andname-expression behaves rather different:
As it refers to only one conditional equation, the order of appearance of arguments is 
preserved in the condition list.

\item
A ``sequential'' {\tt (\seqorname\ L\math{_1} \ldots\ L\math{_n})} is also placed to the specifiers
disposal. This expression guarantees, that all arguments from L\math{_0} to L\math{_{i-1}} are not fulfilled
when the validity of L\math{_i} is checked. To illustrate the difference between {\tt \orname}
and {\tt \seqorname} a characteristic function is specified. It tests, whether all elements in a
list are equal. Here we assume
 {\samepage \tt (car (cons x l)) $=$ x}, \hspace{0.2cm}
 {\samepage \tt (cdr nil) $=$ nil} (!)   \hspace{0.2cm} and
 {\samepage \tt (cdr (cons x l)) $=$ l}.
The specification of {\tt car} need not necessarily be complete.

   \begin{program}
   \pushtabs
   --\= \kill
   (\macrorulename\ (equal-l l) \\
   \>(\ifname\ \=((\seqorname\  \= (= (cdr l) nil) \\
   \>          \>               \> (\andname\ \= (equal-l  (cdr l)) \\
   \>          \>               \>            \> (= \=(car l) \\
   \>          \>               \>            \>    \>(car (cdr l)))))) \\
   \> \> true \\
   \> \> false)) \\
   \poptabs
   \end{program}

The corresponding conditional equations for the {\tt true}-case are:

{\tt 
\begin{tabular}{lclcl}
  (equal-l l)  &\boldequal  &true  &\rulesugar  &(cdr l)           \boldequal\ nil \\
  (equal-l l)  &\boldequal  &true  &\rulesugar  &(cdr l)           $\neq$      nil, \\
               &            &      &            &(equal-l (cdr l)) \boldequal\ true, \\
               &            &      &            &(car l)           \boldequal\ (car (cdr l))\\

\end{tabular}\\
}

The condition list of the second equation contains the negated first  argument of \seqorname\
besides the second one. If an \orname -expression were used in spite of the \seqorname\,
a termination problem would occur because this first negated condition would be removed:

{\tt 
\begin{tabular}{lclcl}
  (equal-l l)  &\boldequal  &true  &\rulesugar  &(equal-l (cdr l)) \boldequal\ true, \\
               &            &      &            &(car l)           \boldequal\ (car (cdr l)) \\
\end{tabular}\\
}

\vfill\pagebreak

{\samepage
As the dual of \seqorname, an \seqandname-expression is also included. 
The \andname- and the \seqandname-expression are
equivalent with respect to the positive/negative-conditional rules they denote
unless its negation becomes relevant due to an outer ``\notname'' or a following \elsename-case.
The \seqandname-expression should be used whenever the order of 
appearance of the arguments is relevant.
}

For an \seqandname-condition  with  \bigmath{n+1} arguments the \ifname-expression saves 
\ \mbox{\math{(n\tight+1)\,\tight*\,(n\tight+2)\,\tight/\,2}} \ 
condition literals
and \math{n} repetitions of  the meta-term of the else-case in the 
specification:
   {\tt \begin{tabbing}
   ---\=---\=---\=  \kill
   (\macrorulename\ l\\
   \> (\ifname\ ((\seqandname\ L\math{_0} \ldots\ L\math{_n})) \\
   \> \> r\math{_0}\\
   \> \> r\math{_1}))
   \end{tabbing}}
written in form of unstructured conditional equations is much longer:

{\tt 
\begin{tabular}{lclcl}
l
&\boldequal
&r\math{_0}
&\rulesugar 
&L\math{_0} \ldots\ L\math{_n}\\
l
&\boldequal
&r\math{_1}
&\rulesugar 
&(not L\math{_0})\\
l
&\boldequal
&r\math{_1}
&\rulesugar 
&L\math{_0} (not L\math{_1})\\
\math\vdots
&&\math\vdots
&
&\math\vdots\\
l
&\boldequal
&r\math{_1}
&\rulesugar 
&L\math{_0} \ldots\ L\math{_{n-1}} (not L\math{_n})\\
\end{tabular}\\
}

For a \casename-with-\elsename-expression 
the saving has the complexity of the 
product of the squares of the numbers of arguments of the \seqandname-expressions.

\end{itemize}

\yestop
\yestop
\yestop
\yestop
\vfill

\noindent 
We now give a final version of our introducing \deletenameintext-specification:
\begin{program}
(macro-rule (delete x l)\\
\>(case
\\
\\\>\>(\match{l}{nil})
\\\>\>nil
\\
\\\>\>(\match{l}{(cons y k)}
\\\>\>~\letc{(delete x k)}{h})
\\\>\>(if ((= x y))
\\\>\>\>h
\\\>\>\>(cons y h))
\\
\\\>\>((not (memberp x l)))
\\\>\>l))
\end{program}
The last case really should be omitted. It is only present to remind
the cursory reader
that the cases must be neither complementary nor complete and that their
ordering is 
(in contrast to LISP's {\tt COND})
relevant only for the order of the tests of an
optional \elsename-case of the \casename-expression.

\vfill\pagebreak

\noindent
All in all,
this \macroruleintext-construct
was designed as a tool for the specifier.
Besides that, it is also useful for explicitly structuring an equational
specification.
This structuring must be done anyway:
\begin{itemize}

\item
It reduces the number of matching and condition tests and therefore
enhances efficiency of rewriting.

\item
More important for us is
that it may exhibit the recursive
construction of a function and therefore may help to find 
suitable structures for inductive proofs
by giving hints for case distinctions and for the choice of covering sets 
of substitutions:

\begin{sloppypar}
For example, the ``natural'' way of proving 
inductive properties of the \deletenameintext-
function
is to start with a covering set of substitutions given by 
``\math\{\ts l\math\mapsto\ts{nil}\math\}'' and
``\math\{\ts l\math\mapsto\ts{(cons~y~k)}\math\}'', and then 
to make a case distinction 
for the second case  
on 
whether ``\ts x=\ts y'' holds 
or not.
\end{sloppypar}
\end{itemize}

\vfill\pagebreak
\section{Examples}

\yestop
In this section we give some more examples.

\yestop
\yestop
\noindent 
Two specifications of the characteristic function of the member predicate:

\begin{program}
(macro-rule (memberp x l) 
\\\>(case 
\\\>\>(\match{l}{nil}) 
\\\>\>false 
\\\>\>(\match{l}{(cons y m)}) 
\\\>\>(if ((= x y)) 
\\\>\>\> true 
\\\>\>\> (memberp x m))))
\end{program}\notop
denotes

{\tt\begin{tabular}{lllll}
memberp x nil&\boldequal&false&&\\
memberp x cons y m&\boldequal&true&\rulesugar&x \boldequal\ y\\
memberp x cons y m&\boldequal&memberp x m&\rulesugar&x \boldunequal\ y\\
\end{tabular}}

\yestop
\noindent
while

\notop\begin{program}
(macro-rule (memberp x l) 
\\\> (case 
\\\> \> (\match{l}{nil}) 
\\\> \> false 
\\\> \> (\match{l}{(cons y m)}) 
\\\> \> (if ((or (= x y) (memberp x m))) 
\\\> \> \> true 
\\\> \> \> false)))
\end{program}\notop
denotes
 
{\tt\begin{tabular}{lllll}
memberp x nil&\boldequal&false&&\\
memberp x cons y m&\boldequal&true&\rulesugar&x \boldequal\ y\\
memberp x cons y m&\boldequal&true&\rulesugar&memberp x m \boldequal\ true\\
memberp x cons y m&\boldequal&false&\rulesugar
&x \boldunequal\ y,\ \ memberp x m \boldunequal\ true\ .\\
\end{tabular}}

\vfill\pagebreak

\yestop
\yestop
\noindent 
Functions on natural numbers:

\begin{program}
(macro-rule (p x)
\\\>   (case
\\\>\>      (\match x {(s u)})
\\\>\>      u))
\end{program}\notop
denotes

{\tt\begin{tabular}{lllll}
p s u &\boldequal&u&&\\
\end{tabular}}

\yestop
\noindent
which is syntactically more restrictive and operationally more useful than

\notop\begin{program}
(macro-rule (p x)
\\\>   (case
\\\>\>      ((= x (s u)))
\\\>\>      u))
\end{program}\notop
which denotes

{\tt\begin{tabular}{lllll}
p x&\boldequal&u&\rulesugar&x \boldequal\ s u\ .\\
\end{tabular}}

\begin{program}
(macro-rule (max x y)
\\\>   (case
\\\>\>      (\match x 0)
\\\>\>      y
\\\>\>      (\match y 0)
\\\>\>      x
\\\>\>      (\match x{(s u)}
\\\>\>      ~\match y{(s v)})
\\\>\>      (s (max u v))))
\end{program}

\notop\begin{program}
(macro-rule (+ x y)
\\\>   (case
\\\>\>      (\match x 0)
\\\>\>      y
\\\>\>      (\match x{(s u)})
\\\>\>      (s (+ u y))))
\end{program}

\notop\begin{program}
(macro-rule (* x y)
\\\>   (case
\\\>\>      (\match x 0)
\\\>\>      0
\\\>\>      (\match x{(s u)})
\\\>\>      (+ y (* u y))))
\end{program}

\notop\begin{program}
(macro-rule (pot w x) 
\>\>\>\>\>\>\>\>; computes w\math{^{\ts x}}
\\\>   (case
\\\>\>      (\match x 0)
\\\>\>      (s 0)
\\\>\>      (\match x{(s u)})
\\\>\>      (* \>w
\\\>\>         \>(pot w u))))
\end{program}

\vfill\pagebreak

\yestop
\yestop
\noindent 
Functions on binary trees:

\begin{program}
(macro-rule (hight t)
\\\>   (case
\\\>\>      (\match t{nil})
\\\>\>      0
\\\>\>      (\match t{(mk-tree l node r)})
\\\>\>      (s (max \>\>\>(hight l) 
\\\>\>              \>\>\>(hight r)))))
\end{program}

\begin{program}
(macro-rule (count-nodes t)
\\\>   (case 
\\\>\>      (\match t{nil})
\\\>\>      0
\\\>\>      (\match t{(mk-tree l node r)})
\\\>\>      (s (+ \>\>(count-nodes l)
\\\>\>            \>\>(count-nodes r)))))
\end{program}

\begin{program}
(macro-rule (completep t)
\\\>   (case
\\\>\>      (\match t{nil})
\\\>\>      true
\\\>\>      (\match t{(mk-tree r node l)})
\\\>\>      (if \>\>((= (hight l) (hight r))
\>\>\>\>\>\>\>\>; |~~this is a conjunctive condition
\\\>\>          \>\>~(completep l)          
\>\>\>\>\>\>\>\>; |~~list, just like with equational
\\\>\>          \>\>~(completep r))         
\>\>\>\>\>\>\>\>; |~~rules
\\\>\>          \>true    \>                   
\\\>\>          \>false)))\>
\end{program}
\vfill\pagebreak

\section{Syntax}

\vfill
\begin{sloppypar}
The syntax of 
the \macroruleintext-construct is defined by the following context-free 
grammar%
\footnote{Here, 
  ``\reps{ ...}~'' 
  denotes zero or more repetitions,
  ``\repsoom{ ...}~'' 
  denotes one  or more repetitions, 
  ``\mbox{...$|$...}''  
  denotes different possibilities,
  ``\nts{...}'' 
  denotes non-terminals, and
  typewriter font indicates grammar terminals.
} 
with starting symbol \macrorule\@.
Note that the sets of variable, constant, and function names
must be mutually disjoint. Furthermore, function names must be 
different from 
``\casename'', and ``\ifname''
and should%
\footnote{This is necessary
  if the function is specified as characteristic function and
  used in a predicate-atom.
}
also be different from
``\ts{=}'',
``\ts{\#}'',
``\ts{def}\,'',
``\matchname'',
``\doublematchname'',
``\letname'',
``\orname'',
``\seqorname'',
``\andname'',
``\seqandname'',
and
``\ts{not}''\@.
\end{sloppypar}

\vfill
\vfill

{\noindent \sloppy 
\begin{tabular}{@{}r r l@{}}
\\
\term               &:= & \variablename \\
                    &$|$& \constantname \\
                    &$|$& \tpar{\functionname~\repsoom{\term}} \vspace{1cm}\\
\inequalityatom     &:= & \tpar{\ts{=} \ \term\ \term} \\
                    &$|$& \tpar{\ts{\#} \ \term\ \term} \vspace{2mm} \\
\predatom           &:= & \term \vspace{2mm}\\
\negatableatom      &:= & \inequalityatom \\
                    &$|$& \predatom \vspace{2mm}\\
\defatom            &:= & \tpar{\ts{def} \ \term} \vspace{2mm}\\
\basicatom          &:= & \negatableatom \\
                    &$|$&\defatom \vspace{2mm}\\
\matchatom          &:= &\tpar{\ts{\matchname     ~~}\ \variablename\ \term}\\
                    &$|$&\tpar{\ts{\doublematchname~}\ \variablename\ \term}
\vspace{2mm}\\
\letatom            &:= &\tpar{\ts{let}\ \term\ \variablename}\vspace{1cm}\\
\negatablecondition &:= & \negatableatom \\
                    &$|$& \tpar{\andname    {\tt~~}  \reps{\negatablecondition}} \\
                    &$|$& \tpar{\orname     {\tt~~~} \reps{\negatablecondition}} \\
                    &$|$& \tpar{\seqandname {\tt~}   \reps{\negatablecondition}} \\
                    &$|$& \tpar{\seqorname  {\tt~~}  \reps{\negatablecondition}} \\
                    &$|$& \tpar{\ts{not~~}           \negatablecondition} \vspace{2mm}\\
\end{tabular}
\vfill\pagebreak                  

\begin{tabular}{@{}r r l@{}}
\generalcondition   &:= & \negatablecondition \\
                    &$|$& \basicatom\\
                    &$|$& \matchatom\\
                    &$|$& \letatom\\
                    &$|$& \tpar{\ts{and~~}\ \reps{\generalcondition}} \\
                    &$|$& \tpar{\ts{or~~~}\ \reps{\generalcondition}} \\
                    &$|$& \tpar{\seqandname{\tt~}  \reps{\negatablecondition}
                          \generalcondition}     \\
                    &$|$& \tpar{\seqorname {\tt~~} \reps{\negatablecondition} 
                          \generalcondition}     
\vspace{2mm}\\
\negatableconditionlist   &:= & \tpar{\reps{\negatablecondition}} \vspace{2mm}
\\
\generalconditionlist     &:= & \tpar{\reps{\generalcondition}} \vspace{1cm}
\\ 
\negatablecase            &:= & \negatableconditionlist \\
                          &   & \metaterm \vspace{2mm}\\
\elsecase                 &:= & \elsename \\
                          &   & \metaterm \vspace{2mm}\\
\case                     &:= & \generalconditionlist \\
                          &   & \metaterm \vspace{1cm}
\\
\ifterm                   &:= & \tpar{\ifname\ \ \ \ \negatableconditionlist \\
                          &   & ~~~\metaterm \\
                          &   & ~~~\metaterm} \vspace{2mm}
\\
\casetermwithelse         &:= & \tpar{\ts{case} \\
                          &   & ~~~\reps{\negatablecase} \\
                          &   & ~~~\elsecase} \vspace{2mm}\\
\caseterm                 &:= & \tpar{\ts{case} \\
                          &   & ~~~\repsoom{\case}} \vspace{1cm}
\\
\metaterm                 &:= & \term \\
                          &$|$& \ifterm \\
                          & $|$& \casetermwithelse \\
                          & $|$& \caseterm \vspace{1cm}
\\
\macrorule                & := & \tpar{\ts{macro-rule} \term\ \metaterm} \vspace{1cm}
\end{tabular}
} 
\vfill\pagebreak

\flushbottom
\newlength\oldparindent
\oldparindent=\parindent
\parindent=0pt

\section{Semantics}

The semantics of a sequence of 
\macroruleintext-expressions 
is a positive/negative-conditional rule system.

Let:
 
\begin{tabular}{l@{~~~$\in$~~~}l}
VAR\idx{i}                         &\LG{\nts{variable-name}}\footnotemark\\
TERM\idx{i}                        &\LG{\term} \\
PRED-ATOM\idx{i}                   &\LG{\predatom} \\
N-C\idx i                          &\LG{\negatablecondition} \\
N-C-LIST\idx i                     &\LG{\negatableconditionlist} \\
BASIC-ATOM\idx{i}                  &\LG{\basicatom} \\
MATCH\idx{i}                       &\LG{\matchatom} \\
LET\idx{i}                         &\LG{\letatom} \\
GEN-COND\idx{i}, G-C\idx i         &\LG{\generalcondition} \\
CASE\idx{i}                        &\LG\case\\
META-TERM\idx{i}                   &\LG\metaterm
\end{tabular}\\
\footnotetext{\LG{\nts{sym}} denotes the set of words generated 
              by productions of our grammar starting from the symbol \nts{sym}.}

\yestop
\yestop
The denotation of the following ``{\em elementary}\/'' 
\macroruleintext-expressions is defined as follows:

{\tt (macro-rule TERM\idx{1} TERM\idx{2})}

denotes the unconditional rewrite rule

TERM\idx{1} ~~~$=$~~~ TERM\idx{2}

and
\begin{program}(macro-rule TERM\idx{1} \\
\> (case \\
\> \> (BASIC-ATOM\idx 0 $\cdots$ BASIC-ATOM\idx{n}) \\
\> \> TERM\idx{2}))
\end{program}
denotes the following rewrite rule with nonempty condition

TERM\idx 1 \math= TERM\idx 2 
~\rulesugar~ BASIC-ATOM\idx 0, \ldots, BASIC-ATOM\idx n 

\yestop
\begin{sloppypar}
A \macroruleintext-expression is non-erroneous \uiff\ it can be transformed 
into elementary \macroruleintext-expressions with the rewrite rules we will 
introduce in this section.
Note that the semantics is declarative in so far as 
no precedence is imposed on the application of these rules.
The resulting rewriting relation is confluent and \noetherian%
.
Since all elementary \macroruleintext-expressions are irreducible,
each \macroruleintext-expression denotes at most one 
positive/negative-conditional rule system.
\end{sloppypar}

\vfill\pagebreak

\subsubsection*{``Predicate''-Removal}

\notop
Predicates may be used as conditions. All these predicates are put into 
equations:

In the context of a general or negatible condition: \\ 
\mbox{} {\tt PRED-ATOM} \mbox{} \redsimple\ \mbox{} {\tt (= PRED-ATOM true)}

\yestop
\subsubsection*{``\ifname''-Removal}

\notop
\ifname-expressions are replaced by ``\casename-with-\elsename''-expressions:

\notop
\begin{program}
(\ifname\ N-C-LIST 
\>\>\>\>\>\>\>\>\>\>(\casename
\\
\> META-TERM\idx{1} 
\>\>\>\>\>\>\redsimple\>\>\>\>N-C-LIST
\\
\> META-TERM\idx{2})
\>\>\>\>\>\>\>\>\>\>META-TERM\idx{1}
\\
\>\>\>\>\>\>\>\>\>\>\>\elsename
\\
\>\>\>\>\>\>\>\>\>\>\>META-TERM\idx{2})
\end{program} 
\notop

\subsubsection*{``\elsename''-Removal}

\notop
As \elsename-statements may cause trouble when replacing 
``\casename\ in \casename'' 
(\cf\ below), they 
must be eliminated before:

\notop
\begin{program}
(case 
\\\> (N-C\idxbbox{1,1} $\cdots$ N-C\idxBox{1,n_1}) 
\>\>\>\>\>\>\>\>\>\>\>\>META-TERM\idx{1} 
\\\>\math\vdots 
\>\>\>\>\>\>\>\>\>\>\>\>\math\vdots 
\\\>(N-C\idxbbox{m,1} $\cdots$ N-C\idxBox{m,n_m}) 
\>\>\>\>\>\>\>\>\>\>\>\>META-TERM\idx{m} 
\\\>\elsename 
\>\>\>\>\>\>\>\>\>\>\>\>META-TERM\idx{m+1})
\end{program}\notop
\math\downarrow
\notop\begin{program}
(case \\
\> (N-C\idxbbox{1,1} $\cdots$ N-C\idxBox{1,n_1}) 
\>\>\>\>\>\>\>\>\>\>\>\>META-TERM\idx{1}
\\\>\math\vdots 
\>\>\>\>\>\>\>\>\>\>\>\>\math\vdots 
\\\> (N-C\idxbbox{m,1} $\cdots$ N-C\idxBox{m,n_m}) 
\>\>\>\>\>\>\>\>\>\>\>\>META-TERM\idx{m}
\\\>((or (not N-C\idx{1,1}) 
\\
\>~.~~~(not N-C\idx{1,2})
\\
\>~.~~~\math\vdots\
\\
\>~.~~~(not N-C\idx{1,n_1}))  
\\
\>~\math\vdots
\\
\>~(or (not N-C\idx{m,1}) 
\\
\>~~~~~(not N-C\idx{m,2})
\\
\>~~~~~\math\vdots\
\\
\>~~~~~(not N-C\idx{m,n_m})))  
\>\>\>\>\>\>\>\>\>\>\>\>META-TERM\idx{m+1})
\end{program}

\begin{sloppypar}
If none of the preceding rewrite rules applies anymore,
then all \negatablename\ atoms are \mbox{(in-)equality} atoms 
and no \ifname-  or \elsename-expressions occur in the specification.
\end{sloppypar}

\vfill\pagebreak

\yestop
\subsubsection*{``\notname''-Removal}

{\tt
\begin{program}%
(not (not N-C))
\>\>\>\>\>\>\>\>\>\redsimple\>\>N-C
\\\\
(not (and~ N-C\idx{1} $\cdots$ N-C\idx{n})) 
\>\>\>\>\>\>\>\>\>\redsimple\>\> 
(or~~ (not N-C\idx{1}) $\cdots$ (not N-C\idx{n}))
\\\\
(not (\seqandname\ N-C\idx{1} $\cdots$ N-C\idx{n})) 
\>\>\>\>\>\>\>\>\>\redsimple\>\> 
(\seqorname~ (not N-C\idx{1}) $\cdots$ (not N-C\idx{n}))
\\\\
(not (or~~ N-C\idx{1} $\cdots$ N-C\idx{n}))
\>\>\>\>\>\>\>\>\>\redsimple\>\> 
(and~ (not N-C\idx{1}) $\cdots$ (not N-C\idx{n}))
\\\\
(not (\seqorname~ N-C\idx{1} $\cdots$ N-C\idx{n}))
\>\>\>\>\>\>\>\>\>\redsimple\>\> 
(\seqandname\ (not N-C\idx{1}) $\cdots$ (not N-C\idx{n}))
\\\\
(not (=  TERM\idx{1} TERM\idx{2})) 
\>\>\>\>\>\>\>\>\>\redsimple\>\> 
(\# TERM\idx{1} TERM\idx{2})
\\\\
(not (\# TERM\idx{1} TERM\idx{2}))
\>\>\>\>\>\>\>\>\>\redsimple\>\> 
(=  TERM\idx{1} TERM\idx{2}) \\
\end{program}
\tt}

{\samepage
\subsubsection*{``\orname''-Removal}
\begin{program}%
(case 
\\\> CASE\idx{1} 
\\\> $\vdots$ 
\\\> CASE\idx n 
\\\> (G-C\idxBox{1}   $\cdots$ G-C\idx{p} 
\\\> ~(or GEN-COND\idx{1} $\cdots$ GEN-COND\idx{r}) 
\\\> ~G-C\idxBox{p+1} $\cdots$ G-C\idx{p+q}) 
\\\> META-TERM 
\\\> CASE\idx{n+1} 
\\\> $\vdots$ 
\\\> CASE\idx{n+m})
\\
\\\math\downarrow
\\\\
(case 
\\\>CASE\idx{1} 
\\\>$\vdots$ 
\\\>CASE\idx{n} 
\\\>(G-C\idx{1} $\cdots$ G-C\idx{p} 
     GEN-COND\idxbox{1} 
     G-C\idx{p+1} $\cdots$ G-C\idxBox{p+q}) META-TERM 
\\\>\math\vdots
\>\>\>\>\>\>\>\>\>\>\>\>\math\vdots
\\\>(G-C\idx{1} $\cdots$ G-C\idx{p} 
     GEN-COND\idxbox{r} 
     G-C\idx{p+1} $\cdots$ G-C\idxBox{p+q}) META-TERM 
\\\>CASE\idx{n+1} 
\\\>$\vdots$ 
\\\>CASE\idx{n+m})
\end{program}

Note that for application of this rule 
no `\elsename' may occur in the \casename-expression.
}

\vfill\pagebreak

\yestop
\subsubsection*{``\seqorname''-Removal}
\begin{program}%
(case 
\\\> CASE\idx{1} 
\\\> $\vdots$ 
\\\> CASE\idx n 
\\\> (G-C\idxBox{1}   $\cdots$ G-C\idx{p} 
\\\> ~(\seqorname\ N-C\idx{1} $\cdots$ N-C\idx{r}) 
\\\> ~G-C\idxBox{p+1} $\cdots$ G-C\idx{p+q}) 
\\\> META-TERM 
\\\> CASE\idx{n+1} 
\\\> $\vdots$ 
\\\> CASE\idx{n+m})
\\
\\\math\downarrow
\\\\
(case 
\\\>CASE\idx{1} 
\\\>$\vdots$ 
\\\>CASE\idx{n} 
\\\>(G-C\idxBox{1} $\cdots$ G-C\idx{p} 
    \\\> ~N-C\idxbox{1} 
\\\>~G-C\idxBox{p+1} $\cdots$ G-C\idx{p+q})
\\\>META-TERM 
\\\>(G-C\idxBox{1} $\cdots$ G-C\idx{p} 
    \\\> ~(not N-C\idxbox{1}) N-C\idxbox{2} 
\\\>~G-C\idxBox{p+1} $\cdots$ G-C\idx{p+q})
\\\>META-TERM 
\\\>\math\vdots
\\\>(G-C\idxBox{1} $\cdots$ G-C\idx{p} 
    \\\> ~(not N-C\idxbox{1}) $\cdots$ (not N-C\idx{r-1}) N-C\idxbox{r} 
\\\>~G-C\idxBox{p+1} $\cdots$ G-C\idx{p+q})
\\\>META-TERM 
\\\>CASE\idx{n+1} 
\\\>$\vdots$ 
\\\>CASE\idx{n+m})
\end{program}
Note that for application of this rule 
no `\elsename' may occur in the \casename-expression.

\vfill\pagebreak

\yestop 
\subsubsection*{``\andname[{\tt *}]''-Removal}
\begin{minipage}{10cm}
\begin{program}%
(case 
\\\> CASE\idx{1} 
\\\> $\vdots$ 
\\\> CASE\idx n 
\\\> (G-C\idxBox{1}   $\cdots$ G-C\idx{p} 
\\\> ~(and{\sf [}*{\sf ]} GEN-COND\idx{1} $\cdots$ GEN-COND\idx{r}) 
\\\> ~G-C\idxBox{p+1} $\cdots$ G-C\idx{p+q}) 
\\\> META-TERM 
\\\> CASE\idx{n+1} 
\\\> $\vdots$ 
\\\> CASE\idx{n+m})
\end{program}
\end{minipage}
~~~~~\redsimple~~~~
\begin{minipage}{5cm}
\begin{program}%
(case 
\\\>CASE\idx{1} 
\\\>$\vdots$ 
\\\>CASE\idx{n} 
\\\>(G-C\idxBox{1} $\cdots$ G-C\idx{p}
\\\>~GEN-COND\idxbox{1} $\cdots$ GEN-COND\idxbox{r} 
\\\>~G-C\idxBox{p+1} $\cdots$ G-C\idx{p+q})
\\\>META-TERM 
\\\>CASE\idx{n+1} 
\\\>$\vdots$ 
\\\>CASE\idx{n+m})
\end{program}
\end{minipage}

\yestop 
\noindent
Note that for application of this rule 
no `\elsename' may occur in the \casename-expression.

\yestop 
\subsubsection*{``\casename-in-\casename''-Removal}
\begin{program}
(case \\
\> CASE\idx{1} \\
\> $\vdots$ \\
\> CASE\idx{m} \\
\> (G-C\idx{1} $\cdots$ G-C\idx{p}) \\
\> (case 
\\\>\>(GEN-COND\idxbbox{1,1} $\cdots$ GEN-COND\idxbbox{1,q_1} ) 
\>\>\>\>\>\>\>\>\>\>\>META-TERM\idx{1}
\\\>\>\math\vdots 
\>\>\>\>\>\>\>\>\>\>\>\math\vdots
\\\>\>(GEN-COND\idxbbox{r,1} $\cdots$ GEN-COND\idxbbox{r,q_r} ) 
\>\>\>\>\>\>\>\>\>\>\>META-TERM\idx{r})
\\\>CASE\idx{m+1} \\
\> $\vdots$ \\
\> CASE\idx{m+n})
\end{program}
$\downarrow$
\begin{program}
(case \\
\> CASE\idx{1} \\
\> $\vdots$ \\
\> CASE\idx{m} 
\\\>(G-C\idx{1} 
   $\cdots$ G-C\idx{p}  GEN-COND\idxbbox{1,1} 
   $\cdots$ GEN-COND\idxBox{1,q_1}) META-TERM\idx{1} 
\\\>\math\vdots 
\>\>\>\>\>\>\>\>\>\>\>\>\math\vdots
\\\>(G-C\idx{1} 
    $\cdots$ G-C\idx{p} GEN-COND\idxbbox{r,1} 
    $\cdots$ GEN-COND\idxBox{r,q_r}) META-TERM\idx{r} 
\\\>CASE\idx{m+1} \\
\> $\vdots$ \\
\> CASE\idx{m+n})
\end{program}
Note that for application of this rule 
no `\elsename' may occur in any of the two \casename-expressions.

\vfill

\subsubsection*{``\matchname''-Removal}
As we do not want match-atoms in our final rule-system we replace all
occurrences of a match-variable {\tt VAR} preceding 
a match-atom \match{VAR}{TERM} with the match-term {\tt TERM}\@.
If the match-variable does not occur in the match-term, we also have to
replace all occurrences of the match-variable in the scope of the match-atom
with the match-term. 
Let \VAR{\mbox{\tt TERM}} denote the set of variables occurring in {\tt TERM}.

\yestop
If \bigmath{{\tt VAR}\in\VAR{\mbox{\tt TERM}}}, then the specifier should
be warned like:
\\\linenomath{``WARNING: \match{VAR}{TERM} re-binds {\tt VAR}''}
and we reduce:\notop
{\tt\begin{program}%
\>\match{VAR}{TERM}\>\>\>\>\>\redsimple\>\>\doublematch{VAR}{TERM}
\end{program}}
Otherwise we reduce:\notop
{\tt\begin{program}%
\>\match{VAR}{TERM}\>\>\>\>\>\redsimple\>\>\doublematch{VAR}{TERM}\\
\>            \>\>\>\>\>    \>\>\letc{TERM}{VAR}
\end{program}}

\vfill
\subsubsection*{``\doublematchname''-Shift-Left}
\vspace\parsep

In case of
\bigmath{
    \VAR{\mbox{\tt BASIC-ATOM}}
    \cap
    (\VAR{\mbox{\tt TERM}}
     \tightsetminus
     \{\mbox{\tt VAR}\}
    )
  \not=\emptyset
}
the condition list below is erroneous. Otherwise we reduce:

\begin{minipage}{6cm}
{\tt
(G-C\idx{1} \\
\mbox{}~$\vdots$ \\
\mbox{}~G-C\idx{m} \\
\mbox{}~BASIC-ATOM \\
\mbox{}~\doublematch{VAR}{TERM} \\
\mbox{}~G-C\idx{m+1} \\
\mbox{}~$\vdots$ \\
\mbox{}~G-C\idx{m+n})
}
\end{minipage}
$\redsimple$~~~~~~
\begin{minipage}{6cm}
{\tt
(G-C\idx{1} \\
\mbox{}~$\vdots$ \\
\mbox{}~G-C\idx{m} \\
\mbox{}~\doublematch{VAR}{TERM}\\
\mbox{}~BASIC-ATOM\{VAR$\mapsto$TERM\} \\
\mbox{}~G-C\idx{m+1} \\
\mbox{}~$\vdots$ \\
\mbox{}~G-C\idx{m+n})\\
}
\end{minipage}

\vfill

\yestop
\subsubsection*{``\letname''-Shift-Right}
If \bigmath{\mbox{\tt VAR}\in\VAR{\mbox{\tt TERM}}}, then the specifier should
be warned like:
\\\LINEnomath{``WARNING: \letc{TERM}{VAR} re-binds {\tt VAR}''}

The following inference rule is the dual of 
``\doublematchname''-shift-left.\vspace{\parsep}

\begin{minipage}{6cm}
{\tt
(G-C\idx{1} \\
\mbox{}~$\vdots$ \\
\mbox{}~G-C\idx{m} \\
\mbox{}~\letc{TERM}{VAR} \\
\mbox{}~BASIC-ATOM \\
\mbox{}~G-C\idx{m+1} \\
\mbox{}~$\vdots$ \\
\mbox{}~G-C\idx{m+n})
}
\end{minipage}
$\redsimple$~~~~~~
\begin{minipage}{6cm}
{\tt
(G-C\idx{1} \\
\mbox{}~$\vdots$ \\
\mbox{}~G-C\idx{m} \\
\mbox{}~BASIC-ATOM\{VAR$\mapsto$TERM\} \\
\mbox{}~\letc{TERM}{VAR} \\
\mbox{}~G-C\idx{m+1} \\
\mbox{}~$\vdots$ \\
\mbox{}~G-C\idx{m+n})
}
\end{minipage}

\vfill\pagebreak

\subsubsection*{``\letname''-``\doublematchname''-Swap}

This is the only non-trivial rewrite rule.

\yestop
\begin{minipage}{6cm}
{\tt
(G-C\idx{1} \\
\mbox{}~$\vdots$ \\
\mbox{}~G-C\idx{m} \\
\mbox{}~\letc{TERM\idx{1}}{VAR\idx{1}}\\
\mbox{}~\doublematch{VAR\idx{2}}{TERM\idx{2}} \\
\mbox{}~G-C\idx{m+1} \\
\mbox{}~$\vdots$ \\
\mbox{}~G-C\idx{m+n})
}
\end{minipage}
$\redsimple$~~~~~~
\begin{minipage}{6cm}\tt(G-C\idx{1} \\
\mbox{}~$\vdots$ \\
\mbox{}~G-C\idx{m} \\
\mbox{}~<X> \\
\mbox{}~G-C\idx{m+1} \\
\mbox{}~$\vdots$ \\
\mbox{}~G-C\idx{m+n})
\end{minipage} \\

with {\tt <X>} defined as follows:
\begin{description}

\item[{\mbox{\tt VAR}\idxbox{1}} = {\mbox{\tt VAR}\idx{2}}:] 
ERROR. \\
There is no reasonable semantics for this 
unless 
\bigmath{
  \ts{TERM}\idx1\sigma
  =
  \ts{TERM}\idx2\xi\sigma
} 
for some \math\xi\ replacing the variables of 
\bigmath{
  \VAR{\mbox{\ts{TERM\idx1}}}
  \cap
  \VAR{\mbox{\ts{TERM\idx2}}}
}
with new distinct variables and \math\sigma\ being a most general unifier
for \ts{TERM\idx1} and \math{\mbox{\ts{TERM\idx2}}\xi}.\footnote{
  \ts{<X>} = 
  {\tt 
    (\doublematchname\math{^\ast}~\math{
               \domres\sigma{\VAR{\mbox{\tt\tiny TERM\idx1}}}
                                 }) 
    (\letname  \math{^\ast}~\math{
         (\domres{(\xi\sigma)}{\VAR{\mbox{\tt\tiny TERM\idx2}}})^{-1}
                                 }) 
  }
  would correspond to our intention.
  \Eg\ for 
  \begin{tabular}[t]{@{}l}
  \letc{(mt~l~y~l)}k
  \\
  \doublematch k{(mt~h\idx1~(cons~y~m)~h\idx2)}
  \\
  \end{tabular}
  \\
  we would choose
  \bigmath{\xi:=\ \{\ \mbox{\tt y}\,\mapsto\,\mbox{\tt z}\ \}};
  \bigmath{\sigma:=\ \{\ 
           \mbox{\tt y}\,\mapsto\,\mbox{\tt (cons~z~m)}\ ,\ \ 
           \mbox{\tt h}\idx1\,\mapsto\,\mbox{\tt l}    \ ,\ \ 
           \mbox{\tt h}\idx2\,\mapsto\,\mbox{\tt l}    \ \}}
  and get
  \\
  \ts{<X>} = 
  {\tt 
    \doublematch y{(cons~z~m)}
    \letc  l{h\idx1}
    \letc  l{h\idx2}
    \letc  z{y}
    .
  }
  
  However, this definition would destroy the confluence of `\redsimple'\@.
  \Eg\ consider the following condition-list where \ts y is an alias for \ts u:
  \ts{(\match x{(s~u)} 
       \match x{(s~y)})} \redsimple\\
  \ts{(\doublematch x{(s~u)} 
       \letc{(s~u)}x 
       \doublematch x{(s~y)}
       \letc{(s~y)}x)}.
  \\
  The latter condition-list reduces in two ways. 
  First with
  \bigmath{\xi:=\ \{\ \},}
  \bigmath{\sigma:=\ \{\ u\,\mapsto\,y\ \}}:
  \\\redsimple\
  \ts{(\doublematch x{(s~u)} 
       \doublematch u y
       \letc{(s~y)}x)}.
  \\
  Second with
  \bigmath{\xi:=\ \{\ \},}
  \bigmath{\sigma:=\ \{\ y\,\mapsto\,u\ \}}:
  \\\redsimple\
  \ts{(\doublematch x{(s~u)} 
       \letc u y
       \letc{(s~y)}x)}.
  \\
  Now the first version reports an error if \ts y occurs to the left
  while the second does not.
  Furthermore, the first version will use the variable \ts y in its scope
  while the second will use \ts u instead.%
}
This case, however, is too unlikely and not important enough
to give semantics for, since this would make single
pass error checking more difficult.

\item
[{\tt VAR\idxbox{1}} $\in$ 
 \VAR{\mbox{\tt TERM}\idx{2}} $\backslash$ $\{{\mbox{\tt VAR}\idx{2}}\}$:] 
\hfill \\
\ts{<X>} = \doublematch{VAR\idx{2}}{TERM\idx{2}} \\
The \letname-term is removed since \ts{VAR\idx{1}} 
is re-bound by the match-atom.
Often, this will not be the intention of the
specifier. Therefore a warning should be given.

\item
[{\mbox{\tt VAR}\idx{1}}~\mbox{\math{\not\in}}~$\{{\mbox{\tt VAR}\idx{2}}\}$ 
 \tightcup\ \VAR{{\mbox{\tt TERM}\idx{2}}}:] \hfill \\
\ts{<X>} = \parbox[t]{8cm}{\tt \doublematch{VAR\idx{2}}{TERM\idx{2}} \\
           \letc{TERM\idx{1}\{VAR\idx{2}$\mapsto$TERM\idx{2}\}}{VAR\idx{1}}}
\\
This should be the normal case.
\end{description}

Note that errors and warnings (case one and two) can be detected easily 
by one single pass over the specification
before starting the rewriting.
This allows error and warning messages to refer to the original
\macroruleintext-constructs, which is necessary for being understandable
for the specifier.

\vfill\pagebreak

\subsubsection*{Splitting}
\begin{minipage}{6cm}
\begin{program}
(macro-rule TERM~~~~~~~~~~~ \\
\> (case \\
\> \> CASE\idx{1} \\
\> \> $\vdots$ \\
\> \> CASE\idx{n}))
\end{program}
\end{minipage}
$\redsimple$~~~~~~
\begin{minipage}{5cm}
\begin{program}(macro-rule TERM \\
\> (case CASE\idxbox{1})) \\
$\vdots$ \\
(macro-rule TERM \\
\> (case CASE\idxbox{n}))
\end{program}
\end{minipage}

\vfill\vfill

By application of the inference rules introduced above, 
all non-erroneous 
\macroruleintext-expressions
can be transformed into the following form:
\begin{program}%
(macro-rule TERM\idx{1} \\
\> (case (MATCH\idxbox{1} $\cdots$ MATCH\idx{m}\\
\> \> \> ~BASIC-ATOM\idxbox{1} $\cdots$ BASIC-ATOM\idx{p}\\
\> \> \> ~LET\idxbox{1} $\cdots$ LET\idx{n})\\
\> \> \> TERM\idx{2}))
\end{program}
or

{\tt (macro-rule TERM\idx{1} TERM\idx{2})} .

\yestop
The transformation into an elementary \macroruleintext-expression  
is attained by the last three rules.

\vfill

\subsubsection*{``\doublematchname''-removal}

In case of
\bigmath{
    \VAR{\mbox{\tt TERM\idx{1}}}
    \cap
    (\VAR{\mbox{\tt TERM\idx{2}}}
     \tightsetminus
     \{\mbox{\tt VAR}\}
    )
  \not=\emptyset
}
the specification is erroneous. 
\linebreak
Otherwise we reduce:

\begin{program}(macro-rule TERM\idx{1} \\
\> (case \\
\> \> (\doublematch{VAR}{TERM\idx{2}}  G-C\idx{1} $\cdots$ G-C\idx{m}) \\
\> \> META-TERM))
\end{program}
$\downarrow$
\begin{program}(macro-rule TERM\idx{1}\{VAR$\mapsto$TERM\idx{2}\} \\
\> (case \\
\> \> (G-C\idx{1} $\cdots$ G-C\idx{m}) \\
\> \> META-TERM))
\end{program}

\vfill\pagebreak

\yestop
\subsubsection*{``\letname''-removal}
\begin{program}
---\=---\= \kill
(case 
\\\> CASE\idx{1} 
\\\> $\vdots$ 
\\\> CASE\idx{m} 
\\\> (G-C\idx{1} $\cdots$ G-C\idx{m} \letc{TERM\idx{1}}{VAR}) 
\>\>\>\>\>\>\>\>\>\>\> TERM\idx{2} 
\\\> CASE\idx{m+1} 
\\\> $\vdots$ 
\\\> CASE\idx{m+n})
\end{program}
$\downarrow$
\begin{program}%
(case \\
\> CASE\idx{1} 
\\\> $\vdots$ 
\\\> CASE\idx{m} 
\\\> (G-C\idx{1} $\cdots$ G-C\idx{m})
\>\>\>\>\>\>\>\>\>\>\> TERM\idx{2}\{VAR$\mapsto$TERM\idx{1}\} 
\\\> CASE\idx{m+1} 
\\\> $\vdots$ 
\\\> CASE\idx{m+n})
\end{program}

\yestop
\yestop
\subsubsection*{``\casename-with-empty-condition''-Removal}
{\tt (macro-rule TERM\idx{1} (case () TERM\idx{2}))} ~~$\redsimple$~~
{\tt (macro-rule TERM\idx{1} TERM\idx{2})}

\vfill

\parindent = \oldparindent

\flushbottom
\notop

\vfill

\end{document}